\documentclass{article}

\usepackage[final]{neurips_2025}

\usepackage[utf8]{inputenc} % allow utf-8 input
\usepackage[T1]{fontenc}    % use 8-bit T1 fonts
\usepackage{hyperref}       % hyperlinks
\usepackage{url}            % simple URL typesetting
\usepackage{booktabs}       % professional-quality tables
\usepackage{amsfonts}       % blackboard math symbols
\usepackage{nicefrac}       % compact symbols for 1/2, etc.
\usepackage{microtype}      % microtypography
\usepackage{xcolor}         % colors
\usepackage{multirow,makecell,booktabs}
\usepackage{graphicx}
\usepackage{siunitx}
\usepackage{natbib}
\usepackage{amsmath}
\usepackage{enumitem}
\usepackage{shortbold}
\usepackage{verbatim}
\usepackage{subcaption}
\usepackage{algorithm}
\usepackage{algorithmic}

\newcommand{\ram}[1] {\textcolor{purple}{RM. #1}}
\newcommand{\fpr}[1] {\textcolor{green}{FPR. #1}}

\newcommand{\E} {{\mathbb E}}

\newcommand{\T} {{\mathbb T}}

\newcommand{\wrappedcell}[2][5cm]{%
  \begin{tabular}{@{}p{#1}@{}}%
  \parbox{#1}{\centering #2}%
  \end{tabular}%
}

\title{
Predicting partially observable dynamical systems via diffusion models with a multiscale inference scheme
}

\author{
\textbf{
\wrappedcell[13cm]{
  Rudy Morel\thanks{Contact: \texttt{rmorel@flatironinstitute.org}}
  \,$^{,1}$, 
  Francesco Pio Ramunno$^{2, 3}$, 
  Jeff Shen$^{4}$, \\
  Alberto Bietti$^{1}$,
  Kyunghyun Cho$^{5}$,
  Miles Cranmer$^{6}$,
  Siavash Golkar$^{1,5}$,
  Olexandr Gugnin$^{7}$,
  Geraud Krawezik$^{1}$,
  Tanya Marwah$^{1}$,
  Michael McCabe$^{1,5}$,
  Lucas Meyer$^{1}$,
  Payel Mukhopadhyay$^{6,8}$,
  Ruben Ohana$^{1}$,
  Liam Parker$^{1,8}$,
  Helen Qu$^{1}$, François Rozet$^{9}$,
  K.D. Leka$^{10,11}$,
  François Lanusse$^{1,12}$,
  David Fouhey$^{5}$,
  Shirley Ho$^{1,4,5}$}
}
\\
\vspace{1pt}
\\
\textbf{The Polymathic AI Collaboration}
\\
\vspace{1pt}
\\
\wrappedcell[12cm]{
$^{1}$Flatiron Institute,
$^{2}$University of Geneva,
$^{3}$FHNW,
$^{4}$Princeton University, 
$^{5}$New York University, 
$^{6}$University of Cambridge, 
$^{7}$University of Kyiv,
$^{8}$University of California, Berkeley, 
$^{9}$University of Liège,
$^{10}$NorthWest Research Associates,
$^{11}$Nagoya University,
$^{12}$Université Paris-Saclay, Université Paris Cité, CEA, CNRS, AIM.
}
}

\begin{document}

\maketitle

\begin{abstract}
Conditional diffusion models provide a natural framework for probabilistic prediction of dynamical systems and have been successfully applied to fluid dynamics and weather prediction. However, in many settings, the available information at a given time represents only a small fraction of what is needed to predict future states, either due to measurement uncertainty or because only a small fraction of the state can be observed. This is true for example in solar physics, where we can observe the Sun’s surface and atmosphere, but its evolution is driven by internal processes for which we lack direct measurements. 
In this paper, we tackle the probabilistic prediction of 
% stochastic, partially observable dynamical systems, 
partially observable, long-memory dynamical systems,
with applications to solar dynamics and the evolution of active regions. 
We show that standard inference schemes, such as autoregressive rollouts, fail to capture long-range dependencies in the data, largely because they do not integrate past information effectively. 
To overcome this, we propose a multiscale inference scheme for diffusion models, tailored to physical processes. Our method generates trajectories that are temporally fine-grained near the present and coarser as we move farther away, which enables capturing long-range temporal dependencies without increasing computational cost. 
When 
% this inductive bias is 
integrated into a diffusion model, we show that our inference scheme significantly reduces the bias of the predicted distributions and improves 
% the 
rollout stability.
%   The abstract paragraph should be indented \nicefrac{1}{2}~inch (3~picas) on
%   both the left- and right-hand margins. Use 10~point type, with a vertical
%   spacing (leading) of 11~points.  The word \textbf{Abstract} must be centered,
%   bold, and in point size 12. Two line spaces precede the abstract. The abstract
%   must be limited to one paragraph.
\end{abstract}

\section{Introduction}

% \dfnote{I think that the paper will benefit from a really clear, straightforward argument. There are quite a few things here and I think they just need some reorganization, and making some claims more specific and some claims more general. I think the paper needs a clear slogan that is repeated ad-nauseum throughout the paper to emphasize the contributions. This sloganizing can be done in a full pass once the paper is more complete}

% \ram{Mention somewhere hidden Markov models.}
%Partial observation of a dynamic system
Probabilistic prediction of dynamical systems is at the heart of many challenging tasks in science and engineering.  Diffusion models have recently shown success in probabilistic prediction for physical systems, especially when they are applied to simulated environments~\cite{kohl2023benchmarking} or to settings such as terrestrial weather prediction~\cite{price2023gencast}, where laboratory settings or advanced data assimilation can recover much of the current system state~\cite{hersbach2020era5}.

% Many real cases, however, are confronted with 
Many real systems are \textit{partially observable}, meaning that data is 
missing, unobtainable, or sufficiently noisy such that at any given time there is inadequate information to accurately infer the underlying state of the system. It follows, then, that there is inadequate information to predict its exact evolution.
In these settings, the correct incorporation of past information can help predict future trajectories.

A prime example of such 
% an under-informed
a partially observable 
system is our nearest star. Key components governing the dynamics of the Sun are not directly observable (e.g, the driving forces beneath the visible ``surface''), and what {\it is} observable is only available via remote sensing.  Nonetheless, predicting this particular system's evolution is important due to the potential impact on technology-based sectors of society arising from solar energetic events~\cite{ImplementationPlan2023}.
While domain experts have identified physical descriptors associated with energetic phenomena such as solar flares~\cite{LekaBarnes2007,sharps,nci_aia}, and relevant ML-ready datasets have been curated and published~\cite{galvez2019machine,Angryk2020,aarps}, there does not yet exist a model (physics-based or ML-based) that can predict future states of solar active regions and their magnetic fields across the spatial and temporal scales relevant to significantly improve prediction for these events~\cite{Leka2019a,AsensioRamos2023LRSP}.  

In this paper we study the problem of predicting partially observable dynamical systems with diffusion models~\cite{ho_2020_ddpm}, motivated by the challenging problem of learning solar dynamics from data. 
As a benchmark to encourage community progress on this problem, we assemble an $8.5$TB dataset of $512\times512$ videos of solar regions containing $12$ fields with measurements of the magnetic vector field and the Sun's atmosphere. 
Diffusion models developed for well-observed fluid simulations~\cite{kohl2023benchmarking} or reanalyzed terrestrial weather data~\cite{price2023gencast} typically use an autoregressive inference scheme to generate future predictions, conditioning on only a few past frames (typically two).
For solar dynamics, however, we find that such models struggle to accurately predict the evolution, showing significant deviation from observations over time. 

To address these limitations, we introduce a new multiscale inference scheme based on “multiscale templates”, which provide an efficient way to condition on distant past information without increasing computational cost. These templates enable the generation of distant future time steps while conditioning on fine-grained present information and coarse-grained past times.
A model trained on generating such videos can then be used to generate arbitrarily long trajectories in the future, by combining different multiscale templates. 
Compared to inference schemes such as standard autoregressive rollouts used in the literature~\cite{kohl2023benchmarking,price2023gencast}, our method predicts a distant future time step from past observations in a single call to the diffusion model, avoiding the accumulation of distribution errors. Furthermore, we condition more frequently, and on a larger portion of past observed data.

\noindent\textbf{Contributions.} 
Our key contributions are: 
\textbf{(a)} We introduce a new multiscale inference scheme tailored to partially observable dynamical systems encountered in Physics.
\textbf{(b)} 
%On the challenging problem of solar prediction, 
% As a demonstration, we target solar dynamics prediction;
On the challenging task of solar prediction,
our multiscale inference scheme outperforms standard schemes from the literature on diffusion models for physics and natural videos, reducing prediction bias and instability.
\textbf{(c)} 
To the best of our knowledge, our model is the first multi-modal diffusion model trained to predict high-resolution solar videos; prior work focuses on single modality, low-resolution data (both in time and space).
\textbf{(d)} 
To encourage competition on the challenging problem of solar prediction, we provide a new multi-modal $8.5$TB dataset of $512\times512$ videos capturing solar regions.
% To encourage competition on the challenging problem of solar prediction, we provide a new multi-modal $8.5$TB dataset of ML-friendly $512\times512$ videos of solar regions.
%
Upon publication, our dataset and model will be made publicly available.

\section{Related works}

%\dfnote{I think a lot of these specifics can be kinda merged: Rather than [X] does A, [Y] does B, [Z] does C. We do D, maybe cluster them. Unlike [X,Y] who fail to do D or [Z] who fails in genral, we do D.} 
\noindent\textbf{Diffusion models for predicting dynamical systems.}
Unlike~\cite{lippe2023,pedersen2025thermalizer}, which employ a diffusion model to learn the distribution of individual states in order to refine predictions from a predictor network, our work falls within the scope of modeling the dynamic of the observations. 
Along these lines, \cite{kohl2023benchmarking,cachay2023dyffusion} address highly observable dynamical systems, like fluids governed by the Navier–Stokes equations, where all relevant variables (e.g., velocity, pressure) are accessible. Other works~\cite{price2023gencast} train on data from complex reanalysis of sparse observations (e.g., the ERA5 dataset~\cite{hersbach2020era5}). Full observation or re-analysis is not always feasible. For instance, in solar dynamics, it is challenging to accurately recover surface observations at even moderate scales~\cite[see, e.g.][]{Barnes2023,Caplan2025}, and becomes especially difficult when attempting to infer the state of the Sun’s interior~\cite{RabelloSoares2024,MasakiHotta2024}, energy transfer~\cite{Tilipman2023} or forces acting on the plasma \cite{Borrero2019,Yang2024}, yet this information is key to predicting solar dynamics. 
Thus, while~\cite{kohl2023benchmarking,price2023gencast} see no benefit 
% from more than two past states, additional state information substantially improves results in our setting.
from using more than two past observations, incorporating additional past steps substantially improves results in our setting.
In that sense, our findings align with those of~\cite{ruiz2024benefits} even though they focused on deterministic models. 
% Diffusion models can also be used for data assimilation and prediction from incomplete observations~\cite{rozet2023score,shysheya2024conditional,huang2024diffusionpde}, but this is an orthogonal direction of work, and a dataset of fully observed states from the system must exist to train the model.
Diffusion models can perform data assimilation and prediction from incomplete observations simultaneously~\cite{rozet2023score,shysheya2024conditional,huang2024diffusionpde}, but this requires a dataset of underlying system states to train the model -- an assumption we do not make in this paper.

\noindent\textbf{Inference schemes for diffusion models.}
% Video diffusion models typically operate in pixel~\cite{ho2022video} or latent space~\cite{blattmann2023,he2022_latent,songweige2023}
The standard autoregressive inference scheme for video diffusion~\cite{ho2022video,blattmann2023,he2022_latent,songweige2023,ruhe2024rolling} consists in generating progressively an entire video by sliding a short window.
Beyond this, Flexible Diffusion Models (FDM)~\cite[FDM]{harvey2022flexible} and Masked Conditional Video Diffusion~\cite{voleti2022mcvd} both adopt flexible conditioning strategies and train a single model with a randomized masking. In particular, \cite{harvey2022flexible} introduces two types of inference schemes.
The first, called “long-range,” generates progressively more distant future frames while conditioning only on recent ones, thereby discarding distant past information. 
The second, called “hierarchy-2,” uses a sliding-window with an initial long-range prediction, but it conditions on past information only at the first iteration.
In contrast, our multiscale inference scheme generates videos at multiple scales and conditions on past information across multiple iterations, which is crucial for recovering information in partially observable dynamical systems.
%

% However, these approaches do not dynamically adapt conditioning over time and may still suffer from rollout instabilities, particularly in domains where the connection to past data is essential. In contrast, our method emphasizes adaptive reuse of early context frames, promoting stability and physically plausible rollouts in challenging domains such as solar dynamics (see Figure~\ref{fig:illustration} and Appendix~\ref{app:dataset}).

% Rolling Diffusion~\cite{ruhe2024rolling} uses a sliding window denoising scheme that progressively predicts frames while assigning increasing noise to later timesteps to reflect uncertainty. Although it produces stable rollouts, its local, fixed conditioning window limits its ability to recover  structures driven by long-range dependencies. Our method overcomes this by combining the strengths of MCVD-style masking with a multiscale inference scheme, which prioritizes conditioning on temporally distant context to better model the causal evolution of solar phenomena.

\noindent\textbf{Machine learning for solar physics.} 
% Machine learning models are being applied across many problems in heliophysics 
Machine learning is increasingly used across heliophysics~\cite{
Camporeale2019,AsensioRamos2023LRSP}, in particular for predicting solar energetic events~\cite{Bobra2015,Nishizuka2018,panos2020,pandey2023,francisco2024,li2025}.
However, these approaches typically perform classification based on selected features rather than modeling the temporal evolution of the solar atmosphere. Other works apply ML to enhance data quality~\cite{Broock2022,Jeong2022, supersynthia, Jarolim2025, gugnin2025} or build large-scale pretrained models~\cite{fm_sdo}, but these also do not predict future physical states.
When it comes to predicting future solar trajectories, many works either focus on a single quantity of interest~\cite{Bai_2021,ramunno_2024_13885515,francisco_videodm} or operate on limited spatiotemporal resolutions. 
For example, \cite{ramunno_2024_13885515, solaris} use at least a 4$\times$ spatial downsampling factor and a temporal resolution no finer than 12h.
% Existing work in modeling future solar physical states differs in modality as well as spatio-temporal sampling. Many focus on one quantity of interest like the radial~\cite{Bai_2021} or line of sight component~\cite{ramunno_2024_13885515} of the photospheric field or one filtergram/channel of data (e.g., AIA~\SI{94} {\angstrom} filtergram~\cite{francisco_videodm}). Many methods have limited spatiotemporal sampling  that misses rapid, detailed solar evolution. 
In contrast, our dataset uses multiple modalities (associated to different instruments);
% , including all components of the HMI vector field and three representative AIA channels capturing multiple heights and temperatures; 
is downsampled only $2\times$ spatially, matching the optical resolution of the instrument; and is captured at 1h sampling rate.

\section{Background: Conditional Diffusion models}

% \dfnote{A lot of this is fairly standard, and so it's a re-tread of known results. For NeurIPS, you do need some math, and this helps with getting your daily value of math... so I'd suggest going fairly quickly or somehow making it important for later by explicitly connecting it to later. For variables, I'm not sure if this is common in the diffusion model literature, but if you want, I've included a style file that has macros for bold letters that are not painful to use $\xB$, $\xB_t$; could help and entirely up to you..}

% \dfnote{Here's an alternate suggestion. Can you merge Section 4 Section 5 and present a merged version of the notation? The problem with explaining this setup is that there are going to be a ton of variables and indices: there are timesteps in both diffusion models as well as temporal diffusion. I don't think it's possible to do a clear merged notation that captures everything all at once, but you could somehow cleanly separate them, or basically abstract away diffusion (e.g., we model $p(x_{1:T}|x_{t\le 0})$ via diffusion-based models).}

This section presents the aspects of conditional 
diffusion models~\cite{sohl_2015,ho_2020_ddpm} most relevant to our work.
% \fpr{In this section, we present the core principles of diffusion models \cite{sohl_2015, ho_2020_ddpm}, highlighting the aspects most relevant to our work.} 

\noindent\textbf{Score-based diffusion model.}
% \fpr{
Score-based generative models~\cite{song2019, song2020}, are a class of generative models that learn to sample from complex data distributions by reversing a gradual noising process. These models define a forward diffusion process in which the input data $\xB\in\mathbb{R}^N$ is progressively corrupted by adding Gaussian noise at various noise levels $\sigma_s$
% . Specifically, at time step $t$, the noised data is given by
\begin{equation}
\xB_s = \xB + \sigma_s \epsilonB ~~,~~ \epsilonB \sim \mathcal{N}(0, I_N).
\end{equation}
The resulting distribution over the noisy data is denoted by $p_s(\xB_s)$ and captures how the original data distribution evolves under increasing noise. 
The generative model learns a reverse denoising process which maps a Gaussian distribution to the distribution of the data~\citep{song2019, ANDERSON1982313, Särkkä_Solin_2019}. This can be described as a stochastic differential equation
\begin{equation}
\label{eq:backward-sde}
{\rm d}\xB_s = -\sigma_s^2\nabla \log p_s(\xB_s) {\rm d}s + \sigma_s {\rm d}W_s,
\end{equation}
and involves the score function $\nabla \log p_s(\xB_s)$. 
% \fpr{why do you use $p_t(x_t)$ and not $p(x_t)$}
%
This score can be obtained by solving a denoising task~\cite{ho_2020_ddpm, song2019, song2020, flow_matching, ddim, edmpaper}. Indeed, if we write $D(\xB,s)$ a function that minimizes the $L^2$ loss
\begin{equation}
\label{eq:denoising-loss}
\E_{\xB\sim p_\mathrm{data},\,\epsilonB\sim\mathcal{N}(0,I_N)}
% \left[ \left\| D(\xB_s, s) - \epsilonB \right\|^2 \right]
\left[ \left\| D(\xB_s, s) - \xB \right\|^2 \right]
~~,~~
\mathrm{with}
~~ 
\xB_s = \xB + \sigma_s \epsilonB.
\end{equation}
% \fpr{Indeed, if we write $D_\theta(x_t, t)$ a function that minimizes the $L^2$ loss
% \begin{equation}
% \label{eq:denoising-loss}
% \argmin_{\theta}
% ~
% \E_{x\sim p(x),\,x_t\sim p(x_t | x)}
% \left[ w_t\left\| D_{\theta}(x_t, t) - x \right\|^2_2 \right],
% \end{equation}
% with $w_t$ a weighting factor for regularization \cite{edmpaper}. Then we can show that the score function is actually given by
% $\nabla_\theta \log p_t(x_t) = \left( D_\theta(x_t, t) - x_t \right) / \sigma_t^2$, with $x_t = x + \sigma_t \epsilon$.}
then we can show~\cite{tweedie1947functions,efron2011tweedie,kim2021noise2score,meng2021estimating} that the score is given by
$\nabla \log p_s(\xB_s) = \left( D(\xB_s, s) - \xB_s \right) / \sigma_s^2$.

Therefore, a diffusion model is trained by learning a neural network $D_\theta$ with parameters $\theta$ on the denoising loss~\eqref{eq:denoising-loss}, and sampled by discretizing the reverse process~\eqref{eq:backward-sde}.

\noindent\textbf{Conditional diffusion model.}
In the paper, beyond modeling the distribution $p(\xB)$ of the data, we focus on modeling conditional distributions $p(\xB|\yB)$ where $\xB$ is a trajectory and $\yB$ is a part of the trajectory itself~\cite{voleti2022mcvd,rozet2025lost}.
% \fpr{In many practical applications, we are interested not only in generating data from the marginal distribution $p(x)$, but rather in sampling from a conditional distribution $p(x|y)$, where $y$ is some observed conditioning variable (e.g., a class label, image caption, or partial measurement) \cite{cg, cfg, palette, mcvd}.}
% With the same framework as above, once can learn a model of a conditional distribution $p(x|y)$. 
% In this paper we only consider learning a conditional distribution of $x$ given a part of the signal itself $x$. 
To that end, let $\mB\in\{0,1\}^N$ denote a vector (or \textit{mask}) indicating which parts of the signal $\xB$ are used as conditioning. The conditioning data is written $\mB \odot \xB$, where $\odot$ is the element-wise product. 
As above, the distribution $p(\xB|\mB \odot \xB)$ can be modeled by learning a denoiser to reconstruct the "clean" data $\xB$ from its noised version $\xB_s$ with in addition the information of the conditioning:
\begin{equation}
\label{eq:conditional-denoising-loss}
\E_{\xB\sim p_\mathrm{data},\,\epsilonB\sim\mathcal{N}(0,I_N)}
\left[ \left\| D((1-\mB) \odot \xB_s + \mB \odot \xB, s, \mB) 
- 
% \epsilonB \right\|^2 \right] 
\xB \right\|^2 \right] 
~~,~~
\mathrm{with}
~~ 
\xB_s = \xB + \sigma_s \epsilonB.
\end{equation}
where the mask $\mB$ is fed to the denoiser $D_\theta$ to help differentiate 
between noised data and conditioning data.
This way, the denoiser is trained to retrieve the global noise from the noised
data $\xB_s$ just like Eq.\eqref{eq:denoising-loss}, but with additional 
conditioning clean information $\mB \odot \xB$.

\definecolor{Fig1Purple}{RGB}{78,33,128}
\definecolor{Fig1Red}{RGB}{255,0,0}
\definecolor{Fig1Blue}{RGB}{0,0,255}
\definecolor{Fig1Gray}{RGB}{50,50,50}

\begin{figure}[t]
\centering
\includegraphics[width=\linewidth]{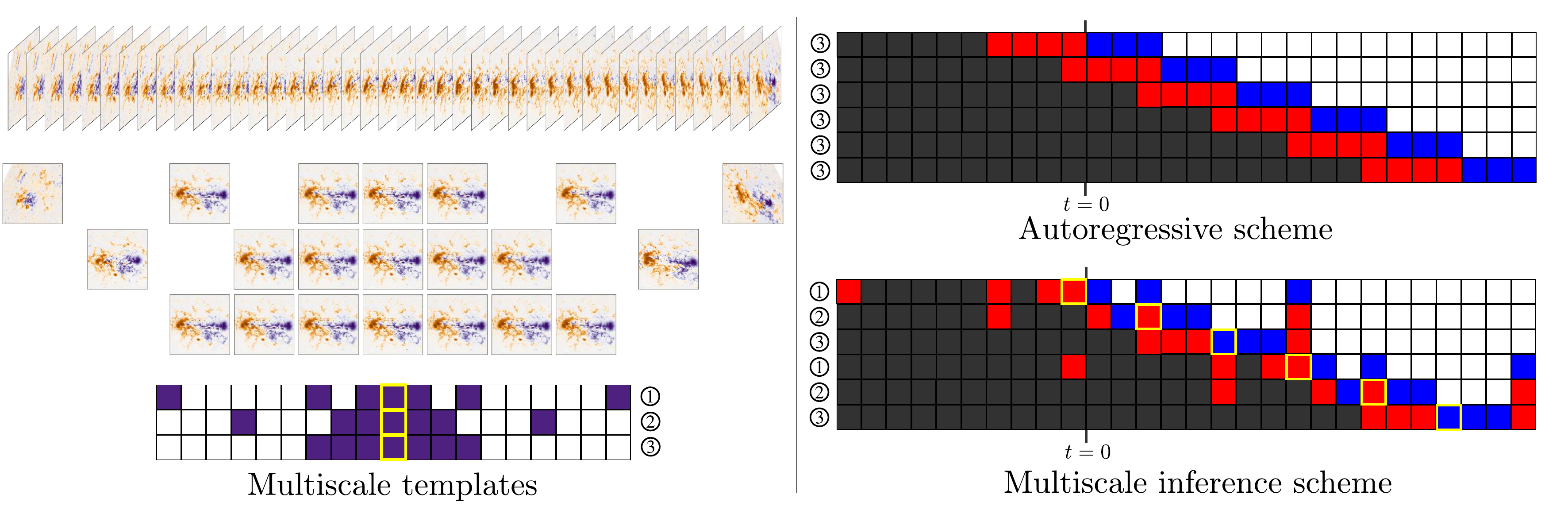}
\caption{
\textbf{Multiscale templates and inference scheme.} 
{\bf (Left)}: Our multiscale templates in \textcolor{Fig1Purple}{\bf purple}. 
{\bf (Right)}: Comparing a standard autoregressive scheme (on top) with our multiscale inference scheme. We use the visualization style of~\cite{harvey2022flexible}, in which \textcolor{Fig1Gray}{\bf dark} boxes indicate available steps (either observed or generated at previous iterations) and \textcolor{Fig1Red}{\bf red} and \textcolor{Fig1Blue}{\bf blue} boxes indicate steps that are used as conditioning or generated, respectively.
Each row is a new call to the conditional diffusion model with the used template indicated by the number next to the row.
Our inference scheme enables capturing longer-range dependencies, conditions more often in the past, and mitigate rollout instability by generating a distant future ($9$ on the figure) in one call to the conditional diffusion model.
}
\label{fig:illustration}
\end{figure}

\section{Multiscale inference scheme for physical processes}
\label{sec:our-framework}

In this paper, we are interested in predicting a dynamical system from its observations $\xB$,
e.g. the magnetic field at the surface of the Sun. 
At each time $t$, we denote $\xB_t$ the observation of the system, which provides only a partial view of the underlying true state.

At present time $t=0$, the goal is to generate a future realization $\xB_{1:T}$ at horizon $T$ conditionally on the past $\xB_{t\leq0}$. 
In doing so we aim to approximate the following conditional distribution
\begin{equation}
\label{eq:full-conditional-distribution}
p(\xB_{1:T} | \xB_{t\leq0})\,.
\end{equation}
Due to computational constraints, modeling the full distribution over long horizons $T$ is infeasible. A common approach is to compress the data to extend the effective context length, as done in latent diffusion models~\cite{blattmann2023,he2022_latent,songweige2023}, but the question remains, how to generate arbitrarily long trajectories using a generative model with a fixed trajectory length?

We assume that our conditional diffusion model can generate only a subset of $2K+1$ time steps at once. 
We seek to use the fixed-size model to produce samples over a far larger set of $T \gg 2K+1$ steps by repeatedly applying the fixed length model.
For convenience, assume that the model always generates $K$ future steps from white noise, and the remaining $K+1$ are conditioning (from the past or present). Generating a trajectory of length $T$ thus requires at least $\lceil T/K \rceil$ steps.
If we define $I_n$ as the set of $K$ new time indices generated and $C_n$ the set of $K+1$ frames used as conditioning, the iterated process amounts to the following approximation:
\begin{equation}
\label{eq:inference-scheme}
p(\xB_{1:T}|\xB_{t\leq0}) 
\approx 
\prod_{n=1}^N p(\xB_{I_n} | \xB_{C_n}).
\end{equation}
A collection of pairs of index sets $(I_n,C_n), 1\leq n\leq N,$ is called an \textit{inference scheme}.
Given the above fixed budget constrain, these sets must satisfy $|C_n|=K+1,|I_n|=K$.
We write $P_n$ the set of indices available at step $n$, which is defined recursively as $P_1 = \{t\leq0\}$ (observed past) and $P_n = P_{n-1}\cup I_n$ (available time steps). 
To properly formalize the problem, we consider inference schemes that satisfy the following properties:
\begin{itemize}
\setlength{\itemsep}{4pt}
\item \textbf{(completeness)} $\cup_{n=1}^N I_n = \{1,\ldots,T\}$
\item \textbf{(admissibility)} $C_n\subset P_n$, the conditioning is done on already generated (or observed) steps
% \item (fixed budget) $|C_n| = K+1$ and $|I_n| = K$
\item \textbf{(efficiency)} $I_k\cap I_\ell = \emptyset$ for $k\ne \ell$, no future step is generated twice 
% \lm{Maybe replace efficient by uniqueness: "Each future state is generated at most once to avoid computational overhead." It might be more efficient to generate multiple times in terms of accuracy if not in terms of computation.} 
% \dfnote{ Efficiency/uniqueness and the rest seem to be different. Completeness and admissible are definitely required; fixed budget is just due to the game you're playing. But efficiency seems qualitatively different. It's definitely a consideration, but it's the type of requirement that'll be catnip for a reviewer. } 
\end{itemize}
For example, an autoregressive inference scheme consists of sliding a fixed-size fine-grained window progressively forward in time,
$C_n = \{(n-1) K,\ldots,nK\}$ and $I_n = \{nK+1,\ldots,(n+1)K\}$ as shown on Fig.~\ref{fig:illustration}. 
This autoregressive inference scheme has several downsides, as evidenced in Tab.~\ref{tab:main-comparison-table} and illustrated in Fig.~\ref{fig:illustration}. The main one being that after the second iteration, there is no explicit conditioning on observed data, which contributes to rollout instability. 
% There are several ways of partially solving this issue, often by creating other issues, which will be discussed in section~\ref{subsec:multicale-inference-scheme}.

% \begin{table}
% \caption{
% \ram{Will certainly go in appendix.}
% }
% \label{tab:schemes-metrics}
% \centering
% \begin{tabular}{lll}
% \toprule
% \multicolumn{2}{c}{Part}                   \\
% \cmidrule(r){1-2}
% Name     & Description     & Size ($\mu$m) \\
% \midrule
% Dendrite & Input terminal  & $\sim$100     \\
% Axon     & Output terminal & $\sim$10      \\
% Soma     & Cell body       & up to $10^6$  \\
% \bottomrule
% \end{tabular}
% \end{table}

% While, several interesting works in the literature of natural videos explored the great variety of possible inference inference schemes~\cite{harvey2022flexible, ruhe2024rolling}, none of them were developed for the class of processes we face in Physics. Tab.~\ref{tab:main-comparison-table} shows metrics of the inference scheme itself. As a matter of fact, it shows that some of the inference schemes developed have downsides by essence when applied to Physics processes, as will be confirmed by experiments in section ~\ref{sec:numerical-experiments}.

\subsection{Multiscale templates for physical processes}
\label{subsec:multiscale-templates}

\begin{figure}[t]
\centering
\begin{minipage}{\linewidth}
\hspace{1.3cm}
\includegraphics[width=0.90\linewidth]{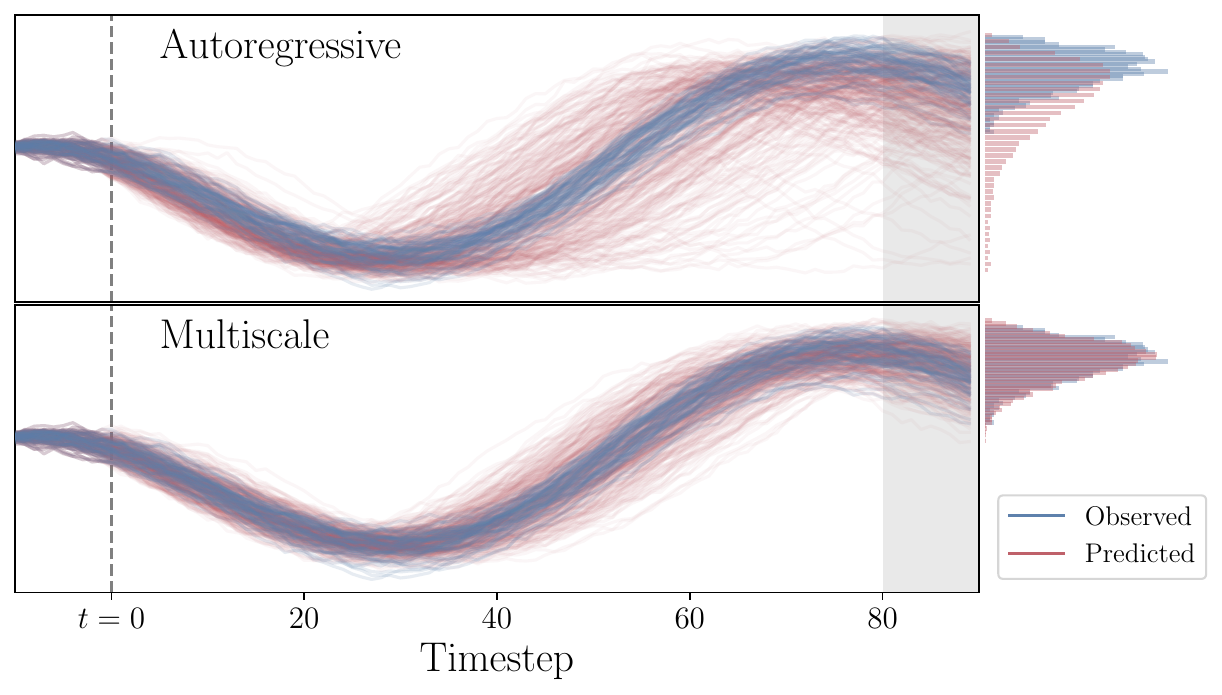}
\end{minipage}
\caption{
\textbf{
Performance of our multiscale inference scheme on a synthetic example.
}
The observed data (blue) consists of Gaussian fluctuations around a sinusoidal trend. Predictions (red) are from a diffusion model with access only to past data $t\leq0$.
\textbf{(Top):} The global trend is barely observable at fine scale. Thus, a model that generates small trajectory segments autoregressively tends to accumulate errors, leading to biased and overly broad predicted distributions.
\textbf{(Bottom):} Our multiscale inference scheme (see Fig.~\ref{fig:illustration}) efficiently recovers the target distribution -- with a Wasserstein distance of 0.021 vs. 0.23 for the autoregressive model. When restricted to the same 3-step past horizon, the multiscale inference still performs better, with a Wasserstein distance of 0.08.
}
\label{fig:synthetic-example}
\end{figure}

Finding an appropriate inference scheme for partially observable dynamical systems is challenging due to the large space of possibilities: many candidates exist for pairs of conditioned times $C_n$ and generated times $I_n$ at each step that satisfy the above properties.

To guide our design, we highlight two key challenges encountered in predicting physical systems:
\begin{enumerate}[label=(\alph*)]
% \item \textbf{Stochastic.} 
% The conditional distribution of the future state given the past cannot be reduced to a Dirac. 
% This uncertainty in the data can originate from chaos, instrumental noise, or microscopic fluctuations. 
\item \textbf{Partially observable.} 
The state of the system at any given time cannot be fully determined from the observations. Consequently, the distribution of future scenarios conditioned on past observed data may not be restricted to a Dirac measure.
% If a system state is knowable and can be described by well-constrained partial differential equations ({\it e.g.}, a magneto-hydrodynamic framework~\cite{Priest1987}), one can in principle solve the dynamics forward in time (Markov process) from a single time step.  
In many cases, the system
state cannot be fully observed due to missing measurements of key physical variables (e.g., 
velocity fields, or unresolved structures), insufficient observational resolution, or corruption arising from instrumental noise.
% These processes are often called hidden Markov processes~\cite{ephraim2002hidden}. In such settings, accumulating observations across different temporal scales can allow the reconstruction of the system's dynamics \cite{takens2006detecting}.
\item 
% \textbf{Slowly decaying time dependencies.}
\textbf{Long-memory.}
Many physical processes exhibit long memory, or long-range dependency, in time.
% , characterized by a smooth decay: states closer to the present have a stronger impact on the future, while the effect of increasingly distant past states gradually diminishes but remains significant. This was well studied in the particular context of scale invariant processes~\cite{mccoy1996wavelet,abry2000wavelets,mandelbrot2013multifractals,morel2025scale}, particularly using wavelets~\cite{stephane1999wavelet}.
This can be quantified by a smooth decay of the autocorrelation (sometimes characterized in the frequency domain by a power-law decay of the power spectrum~\cite{mccoy1996wavelet,stephane1999wavelet,abry2000wavelets,mandelbrot2013multifractals,morel2023compact,morel2025scale}). Intuitively, observations closer to the present have a stronger impact on the future and the influence of distant past observations gradually diminishes while remaining significant.
\end{enumerate}

Diffusion models have been applied to predicting dynamical systems without fully addressing challenge (a) or relying on additional information to overcome it.
% Diffusion models have been applied to prediction on dynamical systems by using additional methods to remove assumption (b).
For example, \cite{kohl2023benchmarking} apply a diffusion model to fully resolved fluids which are effectively Markovian. 
In weather prediction, although the observed data is sparse, data assimilation—also known as reanalysis—enables the reconstruction of missing information, resulting in large datasets of highly informative states~\cite{hersbach2020era5}, on which diffusion models have been successfully trained~\cite{price2023gencast}.
Other models handle missing states, but require clean sates for training~\cite{shysheya2024conditional,huang2024diffusionpde}, which is not always available.

In this paper, we tackle the challenging problem of predicting the observations of a dynamical system presenting the two challenges (a) and (b) simultaneously, as is common across many disciplines. 
For example, in oceanography and climatology, shallow ocean layers are observed while few observations exist for the deep ocean~\cite{Lin2020}; and in seismology, subsurface stress is not directly measured~\cite{subsurfacestree}.
In solar physics, the goal of predicting a future trajectories of active solar regions from available observations (of the magnetic solar surface and hot coronal atmosphere) is challenged by:
% (a) instrumental noise present in the data \cite{hmipipe,hmical}, which is sometimes not fully understood or mitigated \cite{Schuck2016},
% (b) the available data cannot fully describe a crucial component of the state -- in this case, the drivers of the solar magnetic fields and coronal evolution are missed due to the lack of key observations of the interior. 
(a) missing key components of the sate -- in this case, observations of the interior of the Sun, with instrumental noise present in the data \cite{hmipipe,hmical}, which is sometimes not fully understood or mitigated \cite{Schuck2016}.
And (b), the targets that are of predictive interest, e.g., sunspots, have long-range dependencies described by 
plasma diffusion and flow patterns on local, moderate, and global spatial scales \cite{Caplan2025}.  
% ``Surface flux transport'' models based on these processes can forecast large-scale solar dynamical evolution, but they cannot predict new sunspot regions by themselves, and thus cannot forecast the detailed evolution of sunspot groups.  
%Smaller-scale evolution (both temporally and spatially) can be forecast with physical models using data-driven magnetohydrodynamic frameworks that drive the system from estimates of the local forces.  \kdl{ditto re: citations}  They suffer from a similar inability to predict new sunspots appearing, and are generally limited in the horizon window of prediction and can be very sensitive to noise in the input boundary data.

In principle, if the system state was knowable and described by well-constrained partial differential equations ({\it e.g.}, a magneto-hydrodynamic framework~\cite{Priest1987}), one could solve the dynamics forward in time from a single time step (Markov process).
Now, under assumption (a), even if the underlying system is Markovian, its observations may not be predicted deterministically because of the lack of information; such systems are often called hidden Markov~\cite{ephraim2002hidden}). 
The combination of properties (a) and (b) as it is often the case in real cases, encourages a diffusion model to consider not only information near the present but further back in time to access what is needed to predict the future. Inspired by works on long-range temporal processes~\cite{abry2000wavelets,mandelbrot2013multifractals,morel2025scale} and wavelets~\cite{mccoy1996wavelet,stephane1999wavelet,chang2000wavelet,morel2024path}, we introduce a framework to do this.

A \textit{multiscale template} $\mathbb{T}^\alpha_K$ is a set of $2K+1$ indices centered at the present $t^\alpha_0=0$ and becoming progressively coarser farther from it, defined using time increments as powers of $\alpha \geq 1$:
\begin{equation}
\label{eq:multiscale-templates}
\mathbb{T}^\alpha_K = \{t^\alpha_{-K},\ldots,t^\alpha_0,\ldots,t^\alpha_K\}
~~
\text{with}
~~
t^\alpha_{k+1} = t^\alpha_k + \alpha^k
~~
\text{and}
~~
t^\alpha_{-k} = t^\alpha_{-k+1} - \alpha^k
\end{equation}
This set of indices is symmetrical in $t^\alpha_0=0$. 
For $\alpha=1$, we retrieve a standard uniform window used in an autoregressive scheme.
When $\alpha > 1$, the time indices are progressively more spaced as we move away from present. 
We allow $\alpha$ to be real, in that case, the template is mapped to integers through $\T^\alpha_K = \{\text{sign}(t^\alpha_k)\lfloor |t^\alpha_k| \rfloor \ , \ -K\leq k\leq K\}$ where $\lfloor t\rfloor$ is the integer part of $t$. 

For a fixed budget of $K$ times, a multiscale template allows to consider a horizon in the past (and in the future), that is exponential in $K$, while a uniform template $\alpha=1$ has a horizon that is linear in $K$.
As we will see in the next section, this is crucial for capturing long-range dependencies, and helps stabilize long predictions.

The term \textit{template} reflects the flexibility to later separate it into conditioning $C_n$ and newly generated time indices $I_n$ as needed, that is, to apply an arbitrary conditioning mask $\mB$ in Eq.~\eqref{eq:conditional-denoising-loss}. 

\begin{figure}[t]
\centering
% \subfloat{\includegraphics[width=1.0\textwidth]{figs/hmi_aia_data_crop.pdf}} \label{fig:a} \\
% \subfloat{\includegraphics[width=1.0\textwidth]{figs/aia_video_crop.pdf}} \label{fig:b}
\includegraphics[width=\linewidth]{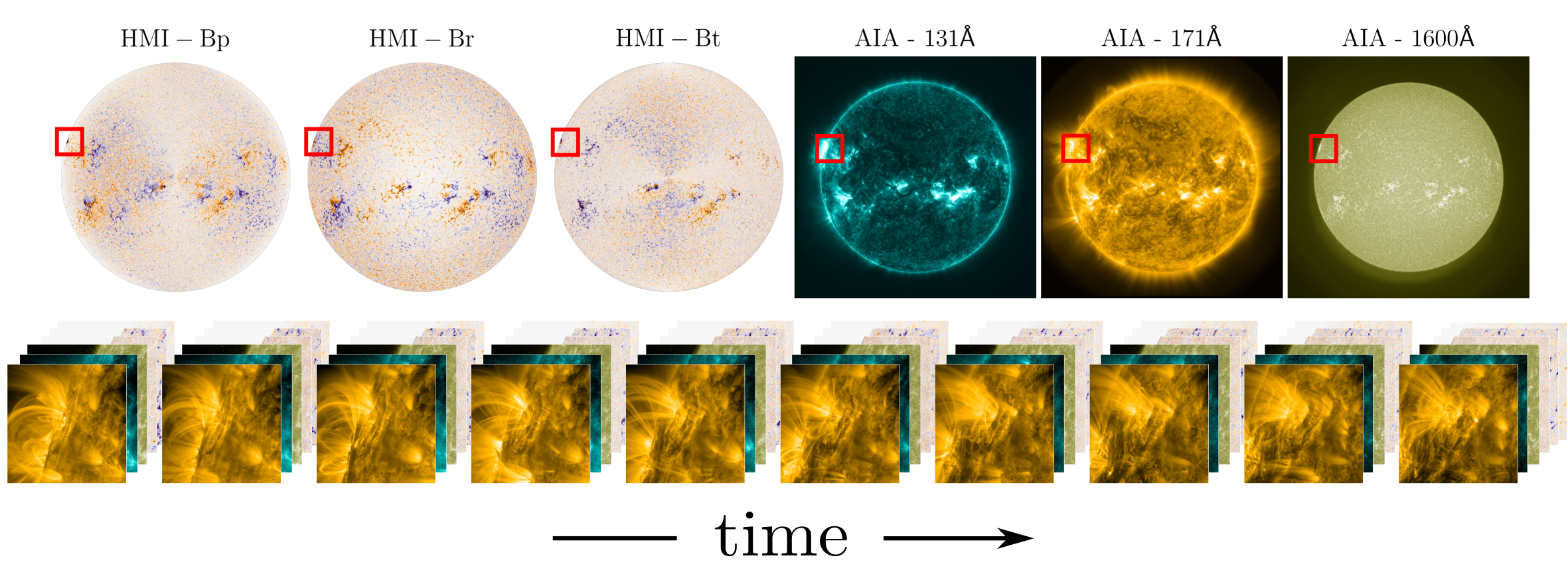}
\caption{
\textbf{(Above):} Example full-disk solar images from 2015-12-12 (see \S\,\ref{sec:solardynamics} for details). The left three panels are photospheric vector magnetic-field components; the right three panels are images of the solar corona and chromosphere. ``Active regions'' (intense magnetic fields connected to bright coronal structures) are present in both modalities. \textbf{(Below):} A sequence of frames of a cropped active region, corresponding to the red box in the row above. 
}
\label{fig:hmi_aia_data}
\end{figure}

\subsection{Multiscale inference scheme}
\label{subsec:multicale-inference-scheme}

\begin{figure}[t]
\centering
\includegraphics[width=1.0\linewidth]{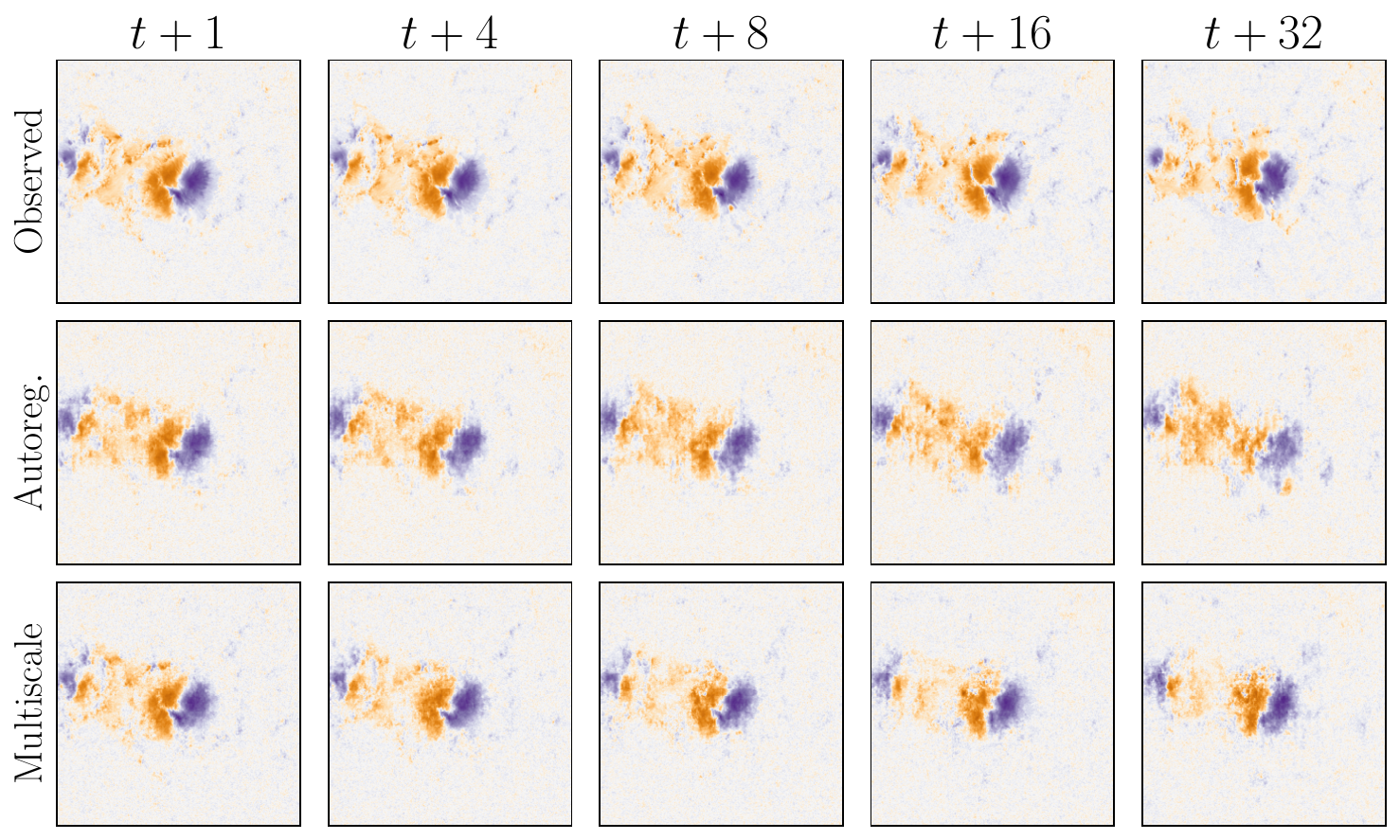}
\caption{
{\bf Example of predictions, for different inference schemes:} 
autoregressive and multiscale (ours).
Colorbar: -3000 \includegraphics[width=24pt,height=7pt]{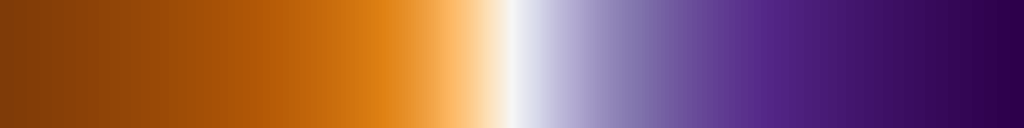} 3000 Gauss (magnetic field).
}
\label{fig:forecasts-trajectories}
\end{figure}

We now design an inference scheme to produce arbitrarily long future trajectories, using the multiscale templates introduced above and motivated by the key properties of observed physical systems.
As described above, this involves defining pairs $(C_n,I_n)$ of conditioning indices and newly generated indices at each iteration $n$, that is, at each call to the diffusion model, which progressively cover a future trajectory (see Eq.~\eqref{eq:inference-scheme}).
In the experiments we choose to generate $K=3$ new time steps at each iteration, which means our diffusion models generate small videos of length $2K+1=7$, and we choose to use templates $\T^{\alpha_\text{max}}_K$ with a maximum $\alpha_\text{max}=2.5$ (see Fig.~\ref{fig:illustration}); in the following we drop the dependence on $K$ and write $\T^\alpha$ directly. 
This means that the most extended video we will generate at once goes up to $9 = \lfloor 1+2.5+2.5^2\rfloor$ steps in the past and future (see Eq.~\eqref{eq:multiscale-templates}). 
We refer the reader to the Appendix for multiscale inference schemes with different choices of $K$ and $\alpha_\text{max}$.

Our inference scheme, illustrated in Fig.~\ref{fig:illustration}, begins by using the largest template $\T^{\alpha_\text{max}} = \{-9,-3,1,0,1,3,9\}$ to generate $K=3$ steps in the future: $I_1 = \{1,3,9\}$ and conditioning on the $K+1=4$ observed steps $C_1=\{-9,-3,-1,0\}$. This enables the model to generate the $9^\text{th}$ step into the future while incorporating observed data that extends equally far into the past. Without completing an entire trajectory, this first step gives us predictions of the physical system at multiple horizons in the future. Once this multiple-horizon prediction is performed, the goal is to "fill the gaps" in the future using the other, shorter-range templates.

Then, we iterate over all possible templates \(\T^\alpha\) with \(1 \leq \alpha \leq \alpha_\text{max}\) in decreasing order, along with all their possible shifts into the future. For each candidate, we check whether the shifted template overlaps with at least \(K+1 = 4\) available time steps. This ensures sufficient conditioning data to generate \(K\) new steps. Among the valid options, we select the first template and shift whose final index aligns with the current maximum horizon, which is 9 in our experiments. 
This ensures that the generation proceeds in a consistent way, gradually filling in missing future steps while maintaining coherence with earlier generated data.
In the experiments, we get $\T = \{-6,-2,-1,0,1,2,6\}$ which must be shifted by $3$ steps in the future. The overlap with the previously generated time steps defines $C_2=\{-3,1,3,9\}$ and the newly generated indices at this second iteration are $I_2=\{2,4,5\}$. 

We repeat this procedure until all the gaps from the first applied largest template are filled. For the values chosen in the experiments, this requires applying a last multiscale template $\T^\alpha = \{-3,-2,-1,0,1,2,3\}$, which is actually a uniform template, shifted by $6$ in the future, and conditioned on the time steps $C_3=\{3,4,5,9\}$ and generating new time steps $I_3=\{6,7,8\}$.

Once the first template span has been entirely generated, we shift the current present to the last generated step, $9$ in the experiments, and can now repeat the above scheme to predict a complete video until $18$ and so on (see Fig.~\ref{fig:illustration}).

This inference scheme offers key advantages. Compared to standard autoregressive or “hierarchy-2” schemes~\cite{harvey2022flexible}, it conditions more often on distant past and future information, better capturing long-range dependencies around the present. It predicts up to 9 steps ahead in a single diffusion call, whereas autoregressive methods require 3 calls for the same horizon. This improves error accumulation, though errors can still grow beyond the largest template’s time scale.

The horizon of the largest template is chosen to be $9$ in experiments but it can be adjusted (see Appendix for a general algorithm). If the physical process exhibits a finite decorrelation timescale, it is natural to choose a largest template that spans this timescale to fully capture long-range dependencies and mitigate rollout instabilities. We refer the reader to the Appendix for multiscale inference schemes based on larger templates.

\section{Numerical experiments}
\label{sec:numerical-experiments}

\begin{table}
\caption{
{\bf Predictions performance.} We compare different inference schemes (Autoregressive, Hierarchy-2~\cite{harvey2022flexible}, Ours -- Multiscale) and models (AViT~\cite{mccabe2024multiple},AR-diff~\cite{kohl2023benchmarking}, Ours). 
For each, we evaluate at three different time windows (1:4 hours, 4:16 hours, 16:32 hours) using multiple metrics: the Wasserstein distance between the distributions; mean absolute error in the power spectrum; and normalized mean absolute error of representative solar physics quantities from~\cite{sharps} -- the Mean Horizontal Gradient of the Total Field 
% $|B|$ 
(MeanGBT) and of the Vertical Field 
% $B_{\rm z}$ 
(MeanGBZ).
}
\vspace{0.2cm}
\label{tab:main-comparison-table}
\centering
\setlength{\tabcolsep}{0.37em}
\newcommand{\smallrange}{{\small 1:4}}
\newcommand{\midrange}{{\small 4:16}}
\newcommand{\bigrange}{{\small 16:32}}
\begin{tabular}{@{}llcccccccccccc}
\toprule
\multicolumn{2}{l}{} & \multicolumn{3}{c}{Wasserstein} & \multicolumn{3}{c}{MAE Power Spec.} & \multicolumn{3}{c}{NMAE MeanGBT} & \multicolumn{3}{c}{NMAE MeanGBZ} \\
\cmidrule(r){1-2} \cmidrule(lr){3-5} \cmidrule(lr){6-8} \cmidrule(lr){9-11} \cmidrule(lr){12-14} 
Model &  Scheme & 
\smallrange & \midrange & \bigrange & 
\smallrange & \midrange & \bigrange & 
\smallrange & \midrange & \bigrange & 
\smallrange & \midrange & \bigrange \\
\midrule
DiT & Autoreg. & 3.9 & 5.6 & 7.9 & 0.25 & 0.36 & 0.53 & 0.18 & 0.30 & 0.37  & 0.15 & 0.25 & 0.31 \\
DiT & Hiera.~\cite{harvey2022flexible} & \textbf{3.0} & 4.6 & 6.0 & \textbf{0.12} & 0.27 & 0.38 & \textbf{0.12} & 0.28 & 0.38 & \textbf{0.09} & 0.22 & 0.31\\
DiT & Ours & \textbf{3.0} & \textbf{4.3} & \textbf{5.5} & \textbf{0.12} & \textbf{0.22} & \textbf{0.33} & 0.14 & \textbf{0.27} & \textbf{0.33} & 0.10 & \textbf{0.21} & \textbf{0.27} \\
\midrule
\cite{mccabe2024multiple}  & Autoreg.  & 12 & 13 & 15 & \textbf{0.11} & 0.35 & 0.81 & 0.40 & 0.44 & 0.45 & 0.40 & 0.43 & 0.44  \\
\cite{kohl2023benchmarking} & Autoreg. & 7.3 & 12 & 16 & 0.20 & 0.47 & 0.71 & 0.29 & 0.52 & 0.67 & 0.27 & 0.49 & 0.64 \\
DiT & Ours & \textbf{3.0} & \textbf{4.3} & \textbf{5.5} & \textbf{0.12} & \textbf{0.22} & \textbf{0.33} & \textbf{0.14} & \textbf{0.27} & \textbf{0.33} & \textbf{0.10} & \textbf{0.21} & \textbf{0.27}
\\
\bottomrule
\end{tabular}
\end{table}

\subsection{Synthetic example}
\label{subsec:synthetic-example}

We present a synthetic example of time-series of observations $x_t=\mu_t+\eta_t,$ where $\mu_t$ is a deterministic sinusoidal trend, and is made partially observable by the addition of Gaussian noise $\eta_t$.
In the absence of noise, a single time step suffices to determine the future trajectory completely. 
In the presence of noise, however, consider the times around a negative peak (approximately $t=30$; see Fig.~\ref{fig:synthetic-example}). Depending on the noise realization, the local trend may be upward or downward, making the state difficult to recover locally. 
That is, partial observability induced by noise prevents accurate estimation of the underlying slowly varying component.
It is thus necessary to look further into the past, which is precisely what our multiscale inference scheme achieves.

Fig.~\ref{fig:synthetic-example} shows predictions with a small diffusion model, with either an autoregressive scheme or our multiscale inference scheme. Our scheme better captures the trajectories than the autoregressive one, as confirmed visually and by Wasserstein distance ($0.021$ vs $0.23$). 
Because of the partial observability of the trend mentioned above, the autoregressive scheme produces errors that accumulate.

%
% \begin{table}[h!]
% \caption{
% {\bf Synthetic example: effect of slow component amplitude.} 
% Performance of our method with varying observability factor $\lambda$ in a synthetic example. 
% We report the Wasserstein distance between model distribution and target distribution (see Fig.~\ref{fig:synthetic-example} for observed trajectories). 
% % Our method 
% }
% \vspace{0.2cm}
% \label{tab:varying-observability}
% \centering
% \setlength{\tabcolsep}{0.5em} % adjust spacing as needed
% \begin{tabular}{@{}lcccc@{}}
% \toprule
% Scheme & \multicolumn{4}{c}{Observability factor $\lambda$ } \\
% \cmidrule(lr){2-5}
%        & 0.0 & 0.1 & 0.2 & 1.0 \\
% \midrule
% Autoregressive       & \textbf{0.055} & 0.13  & 0.19  & 0.26 \\
% Multiscale (ours)    & 0.060 & \textbf{0.027} & \textbf{0.036} & \textbf{0.067} \\
% \bottomrule
% \end{tabular}
% \end{table}
%

Our multiscale scheme efficiently captures long-range dependencies through its multiscale templates (see Section~\ref{subsec:multiscale-templates}). When predicting the future at $t=0$, it also conditions on earlier steps (up to $-9$) compared to only $-3$ for an autoregressive scheme (see Fig.\ref{fig:illustration}).
To isolate the effect of the multiscale template from that of conditioning further in the past, we restrict our scheme in Fig.\ref{fig:synthetic-example} to the same past horizon ($-3$). Performance slightly degrades (from $0.021$ to $0.08$), but still surpasses autoregressive baselines.

We refer the reader to the Appendix for another synthetic example of a partially observable fluid dynamical system.

\subsection{Solar dynamics prediction}
\label{sec:solardynamics}

\noindent\textbf{Solar dataset.}
To encourage competition on predicting partially observable long-memory dynamical systems, we introduce a new ML-ready dataset (see Fig.~\ref{fig:hmi_aia_data}) of reasonably high-resolution solar dynamics prediction based on real observations from the NASA Solar Dynamics Observatory mission~\cite{sdopaper}, in continuous operation since 2010. 
The data contains two modalities from two instruments, surface magnetic fields~\cite{hmipaper}, and images of the solar atmosphere~\cite{aiapaper}.
Each produces $4096\times4096$-pixel images of the full disk of the Sun (see Fig.~\ref{fig:hmi_aia_data}) at high cadence, making the data-handling very demanding.  
% Curation for use with ML requires downsampling in spatial and temporal domains \cite[see discussions in][]{galvez2019machine,Angryk2020}.
%Working on these full-disk images is challenging due to the size and because little of the Sun are active regions. %as the size of the data requires would require compressing it somehow, and most importantly, only a small fraction of what happens on the Sun is of interest to predict. 
As discussed in \cite{aarps}, because active regions occupy only a small fraction of the visible disk, we propose a dataset of square-image videos of $512\times512$-pixel windows that track an active-region. This data is curated to carefully account for the rotation of the Sun, the limb of the Sun (its ``edge''), co-alignment between the two modalities, potential overlap between targets, and uncertainty, artifacts, and missing data.
Each day, we randomly sample $8$ regions of the Sun to follow for $48$h, sampled hourly. The regions are selected to avoid bias towards rare events. 
Our dataset consists of $8.5$TB composed of $\approx15$K multi-channel videos of shape $48\times12\times512\times512$. Each video contains $3$ magnetic fields channels and $9$ channels for the solar atmosphere at different wavelengths.  
In the following, all models are trained on images downsampled by a factor of 2 (to the instrument's optical resolution) and considering only 3 of the atmosphere channels, in order to reduce the computational cost of training multiple diffusion models.

\noindent\textbf{Diffusion model hyperparameters.}
% \ram{Add the positional encodding we do.}
We adopt a Vision Transformer~\cite[ViT]{vit} architecture as our denoiser backbone, following the approach in~\citep{edmpaper}, but extended to handle spatio-temporal data and inspired by the implementation in~\citep{rozet2025lost}. 
The denoiser takes as input 3D patches of size $1 \times 8 \times 8$ (no patchification in time), and consists of $16$ attention-based layers with a hidden dimension of $512$ and $4$ attention heads per layer. The resulting denoiser has $62$ million learnable parameters. Time and spatial information on the patches are added as input and we use a RoPE positional encoding~\cite{su2024roformer}. Like in~\cite{edmpaper}, input, output, and noise levels are preconditioned to improve the training dynamics.
For sampling, we generate small trajectories of length $7$ with $100$ diffusion steps with a Adams-Bashforth multi-step sampler~\cite{wanner1996solving,zhang2022fast}.
% , an integration-based algorithm inspired by the Elucidated Diffusion framework of~\citep{edmpaper}, designed to improve sample quality and stability through adaptive noise scheduling.

% \ram{ Quick description of the denoiser we used, with its hyperparameters, architecture, patch size, number of layers etc ... (model size). Hyperparameters for sampling through diffusion. Number of diffusion steps etc ... }

\noindent\textbf{Evaluation metrics.}
We use several metrics that can be computed between a sample and an observation. In evaluating magnetogram predictions, per-pixel averages are not informative since they are dominated by quiet Sun pixels even in patches~\cite{supersynthia,Higgins21}. We therefore use multiple other metrics (see Tab.~\ref{tab:main-comparison-table}). 
First, the Wasserstein distance assesses the fit between the predicted distribution of pixels and the observed one. 
Second, we compute the mean absolute error in the isotropic power spectrum, which provides information on the spatial frequency content of an image. This metric is less sensitive to noise in the data.
Finally, we consider physics-based summary statistics 
%called SHARPs~\cite{sharps}, which domain experts identified as crucial to predict solar events. We 
%reporting two representative quantities 
that characterize spatial gradients of the magnetic field. 
All metrics are averaged on all fields, on several realizations of the model, at several prediction dates, and averaged over several different time horizons.
% Given we are working on highly stochastic data, standard mean absolute errors or mean square error on the entire image, did not show strong discriminative power.

\noindent\textbf{Baselines.}
We compare our model to $4$ baselines. Two fix the denoising architecture and compare the multiscale inference scheme with: an autoregressive inference scheme (a default choice in the literature) and the hierarchy-2 inference scheme from~\cite{harvey2022flexible} (which sparsely completes missing frames, then autoregressively samples the remainder by conditioning on both past and future frames).
The other two compare our model to existing spatiotemporal models for physical systems: \cite{kohl2023benchmarking} is a diffusion model tested on fluid dynamics data; and \cite{mccabe2024multiple} is a deterministic transformer based on axial attention~\cite{ho2019axial}.
%~\cite{kohl2023benchmarking,mccabe2024multiple}. The first is a diffusion model tested on fluid dynamics data, the second is a deterministic transformer using axial attention~\cite{ho2019axial}.
All models are trained with 40 epochs. We refer the reader to the Appendix for additional details.

\noindent\textbf{Solar predictions.} 
Tab.~\ref{tab:main-comparison-table} confirms that, in this more challenging case, our multiscale inference scheme better predicts the pixel distributions than an autoregressive scheme at all future horizons (1:4, 4:16 and 16:32) by achieving the lowest Wasserstein distance.  
The spatial content is better preserved, shown by the error in the power spectrum, and illustrated in the predictions in Fig.~\ref{fig:forecasts-trajectories}.
Our multiscale inference scheme also outperforms the ``Hierarchy-2'' model introduced for natural videos~\cite{harvey2022flexible}, which was not designed for slow-decaying, autocorrelated long-memory processes.
Tab.~\ref{tab:main-comparison-table} also shows that our diffusion model, equipped with our multiscale inference scheme, significantly outperforms existing models~\cite{mccabe2024multiple,kohl2023benchmarking}. A deterministic baseline such as AViT~\cite{mccabe2024multiple} can predict a future trajectory that is close to observed data but loses high frequency content, which gives rise to errors that accumulate with the rollout. 
Our model also compares favorably to the diffusion model of~\cite{kohl2023benchmarking}, which was developed for fluid dynamics data.
These results showcase the limits of current models in probabilistic prediction of partially observable dynamical systems.

\section{Conclusion and discussion}

This work introduces and analyzes a multiscale inference scheme for predicting partially observable dynamical systems. Our approach efficiently incorporates past information—while being refined around the present—to predict future time steps. We show superior performance in both synthetic settings and the challenging task of predicting solar dynamics, outperforming existing schemes~\cite{harvey2022flexible} and models~\cite{mccabe2024multiple,kohl2023benchmarking} for video and spatiotemporal physical systems prediction. Our results suggest that multiscale temporal conditioning helps mitigate partial observability, especially when long-range precursors influence future evolution, as in solar dynamics. To support further work, we contribute a dataset of high-resolution multi-modal solar regions trajectories. 

% While our approach improves long-horizon coherence without additional computational cost, it relies on pre-defined temporal templates and inference scheme. This limits our method's flexibility in domains with unknown or complex correlation structure. Future work could explore adaptive or learned conditioning schedules.
While our method is well suited for long-memory systems with smoothly decaying temporal dependencies, it may not remain competitive when observations are dominated by short-term patterns. Future work could explore adaptive or learned conditioning strategies.

%%%%%%%%%%%%%%%%%%%%%%%%%%%%%%%%%%%
%DF: I am moving the appendix because it will not be in the final version of the paper that is submitted. I want any citations coming to the appendix to break so we can spot them now, rather than after the deadline etc.
%
%
%%%%%%%%%%%%%%%%%%%%

\appendix

\begin{ack}

The authors thank the Scientific Computing Core at the Flatiron Institute, a division of the Simons Foundation, for providing computational resources and support. They also thank Mark Cheung, Patrick Gallinari, Florentin Guth, and Ruoyu Wang for insightful discussions.

Polymathic AI acknowledges funding from the Simons Foundation and Schmidt Sciences.

\end{ack}

\bibliographystyle{plain}  % or unsrt, ieee, etc.
\bibliography{biblio}

\clearpage

\appendix

\section{Solar dataset}
\label{app:dataset}

\subsection{Full-disk images}

%\ram{Nothing new here, just a quick description of the AWS dataset}

%\ram{Do two paragraphs for the two modalities}
We construct our dataset from full-disk observations captured by NASA's Solar Dynamics Observatory (SDO) \cite{sdopaper}, which has been in continuous operation since 2010. In particular, we use two instruments aboard SDO: the Helioseismic and Magnetic Imager \cite{hmipaper} and the Atmospheric Imaging Assembly \cite{aiapaper}, which together provide a comprehensive view of solar activity across the surface and atmosphere.

\paragraph{Magnetograms (surface of the Sun).} The Helioseismic and Magnetic Imager \cite[HMI]{hmipaper} captures vector magnetograms of the solar photosphere, measuring the magnetic field in three orthogonal components. These observations are acquired at a cadence of 12 minutes and a spatial resolution of 1 arcsecond per pixel, producing $4096 \times 4096$ pixel full-disk images. The magnetic field values span an average dynamic range from $-3000$ to $+3000$ Gauss. 
Since direct measurement of magnetic fields in the solar corona, the Sun's outermost atmospheric layer, HMI magnetograms serve as the primary constraint on the magnetic environment of the outer solar atmosphere. They are thus essential for studying the magnetic drivers of solar activity. The temperature at the photosphere is $\sim 4500$ K. An example of the 3d vector magnetic field captured from HMI is shown in Figure~\ref{fig:hmi_aia_data}. 

\paragraph{Atmospheric images (atmosphere of the Sun)}The Atmospheric Imaging Assembly \cite[AIA]{aiapaper} complements HMI by observing the upper layers of the Sun, ranging from the chromosphere to the outer corona, using multiple ultraviolet (UV) and extreme ultraviolet (EUV) channels. AIA operates at a cadence of 12 seconds with a spatial resolution of approximately 1.5 arcseconds, capturing dynamic atmospheric phenomena across a range of temperatures (from $\sim 10^4$ K to beyond $10^7$ K). These passbands reveal radiative signatures of flares, eruptions, and coronal loops. While AIA does not directly measure magnetic fields, it provides crucial indirect evidence of the coronal response to magnetic activity rooted in the photosphere. A representative set of multi-channel AIA images is shown in Figure~\ref{fig:hmi_aia_data}.

The raw data produced by these instruments is curated in a dataset introduced in~\cite{galvez2019machine}, which includes preprocessing steps such as degradation correction, removal of faulty observations, and temporal co-registration. 
% These corrections ensure that the data are physically consistent and suitable for robust machine learning analysis.

% Only a few regions of the full-disk are of interest, and modeling the time dynamics of the entire full-disk requires downsampling drastically the data~\cite{francesco}. Rather, we propose to 

%Using this archive, we extract hourly observations between 2013 and 2019 from both instruments. For the AIA data, we focus on three representative passbands: \SI{131}{\angstrom}, \SI{171}{\angstrom}, and \SI{1600}{\angstrom}, chosen to span a range of temperature regimes from the photosphere to the flaring corona. This results in a multi-modal observational framework where photospheric magnetic fields (HMI) act as the drivers and atmospheric emissions (AIA) represent structured responses in the corona.

\subsection{Sun regions dataset}
\begin{figure}[t]
\centering
\includegraphics[width=1.0\textwidth]{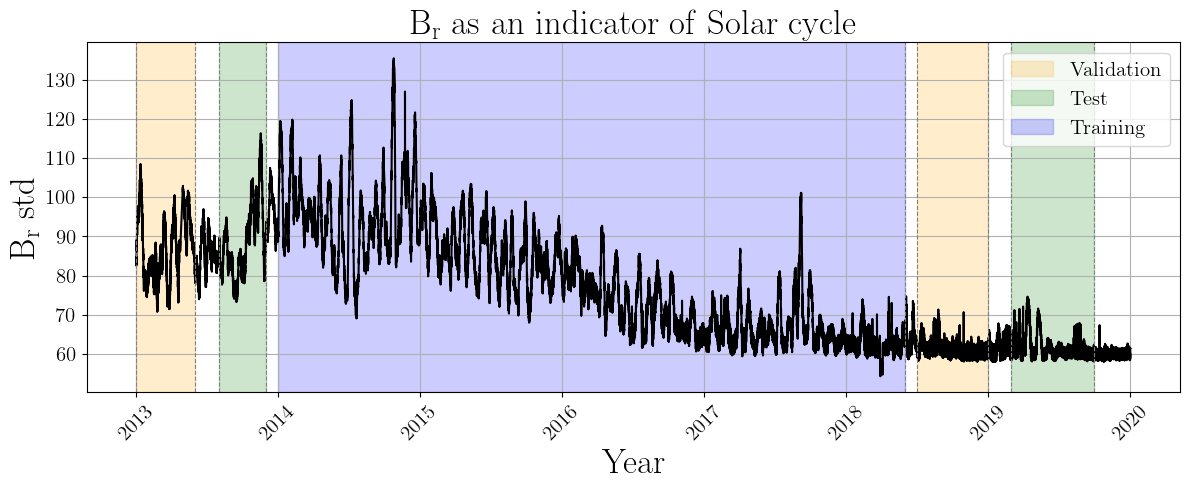}
\caption{
\textbf{Solar activity.}
The standard deviation of the $B_{\rm r}$ component over the full solar disk as a proxy for solar activity, illustrating the temporal segmentation used for model development. 
The dataset is divided into training (blue), validation (yellow), and test (green) intervals, with each evaluation segment separated from training data by at least one full month. 
The split ensures disjoint active regions across sets and includes test intervals during both high and low solar activity, enabling robust model evaluation across varying solar conditions.
}
\label{fig:solar_cycle}
\end{figure}

%\subsection{Sun-region tracking and active-region crop construction} \fpr{I suggest this as a title}
%\ram{The title is too long, rather do paragraphs to describe the steps to assemble the dataset.}
% \ram{The title should tell people that this is the section giving details on the dataset we introduce in this paper: OUR contribution}
% \fpr{what about this: "Crop It Like It’s Hot: Dataset of Solar Active Regions" I mean chatgpt gave me this title based on a Snoop Dogg song and I find it dope ahah}
While full-disk data offer a comprehensive view of the Sun, the majority of solar activity relevant to forecasting tasks is concentrated in localized regions known as active regions. These regions, though occupying only a small fraction of the solar disk, are the primary sources of variability and eruptive events. To concentrate on the most relevant areas and reduce computational cost, we build our dataset from video crops centered on tracked active regions.
\paragraph{Spatial sampling of active regions.} 
From 2013 to 2019, we sample 8 spatial locations per day using a probabilistic strategy based on the absolute value of the radial magnetic field component $|B_r|$ from HMI magnetograms. 
This field component serves as an effective proxy for activity since high $|B_r|$ values correlate strongly with
% magnetic complexity and 
the likelihood of solar eruptions \cite{LekaBarnes2003a, Bobra2015}. 
The sampling probability at each pixel location $x$ is defined as proportional to $\exp(|B_r(x)| / T)$, where $T$ is a tunable temperature-like parameter controlling the sharpness of selection. This prioritizes magnetically active zones while preserving some randomness to avoid bias toward rare extreme events. Regions below a minimum signal-to-noise threshold are excluded, and spatial diversity is enforced via a minimum distance constraint between samples.
\paragraph{Temporal tracking and data extraction}
Once locations are selected, we track their motion across the solar disk using precomputed differential rotation maps provided by the curated dataset of \cite{galvez2019machine}. This tracking compensates for the Sun’s differential rotation, allowing us to follow the same region over time. For each selected location, we extract a $512 \times 512$ pixel crop every hour over a 48-hour window, resulting in a temporally consistent sequence. Each crop includes 12 channels: the three orthogonal components of the HMI magnetic field and nine co-aligned AIA EUV/UV channels. 
% The patches are then downsampled to $256 \times 256$ to reduce the computational load for model training.
% \paragraph{Final dataset format and splitting}
\paragraph{Train/val/test datasets.}
In order to reduce the computational cost of our trainings, the patches are downsampled to $256 \times 256$ and select only $3$ AIA channels: the \SI{131}{\angstrom}, \SI{171}{\angstrom}, and \SI{1600}{\angstrom} bands.
Each resulting sample is a spatiotemporal tensor of shape $48 \times 12 \times 256 \times 256$, representing a 48-hour evolution of solar activity within an active region. We partition the dataset into training (70\%), validation (15\%), and test (15\%) sets. The training set spans January 2014 to May 2018. The validation set comprises two disjoint intervals: January to May 2013 and July 2018 to January 2019. The test set includes data from August to November 2013 and from March to September 2019. To prevent temporal leakage and ensure unbiased evaluation, we insert one-month buffer zones between splits, avoiding overlap of active regions across different subsets. The test periods are deliberately chosen to span both solar maximum and minimum phases (see Figure~\ref{fig:solar_cycle}), supporting a robust assessment of model generalization under diverse solar conditions.

\begin{figure}[t]
\centering
\includegraphics[width=0.8\textwidth]{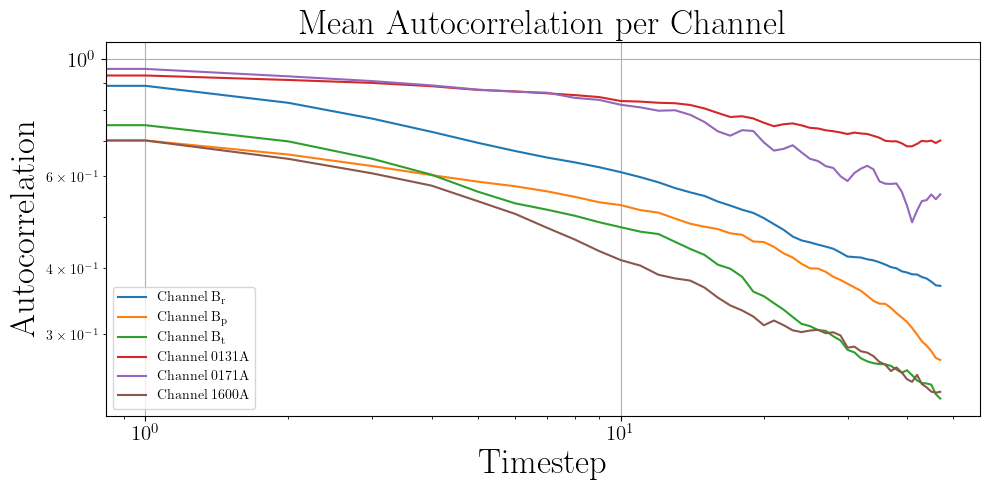}
\caption{
\textbf{Long-memory on solar data.}
Autocorrelation as a function of time for several representative channels in our dataset. All channels exhibit a slow and smooth decay of autocorrelation over time, confirming the presence of persistent long-range dependencies. 
This motivates the use of our multiscale inference scheme, which uses exponentially spaced temporal templates to efficiently capture long-range dependencies.}
\label{fig:app-autocorrelation}
\end{figure}

\section{Additional model description}
\label{app:model}

%\ram{We should add compute ressources: single node of 8 H100 GPUs}
%\ram{hypeparameters should be there: learning rate, batch size etc ... (see config files in the code)}

\subsection{Inference scheme implementation}
%\ram{After the general comment on making the comparison fair, add one paragraph per baseline (auto-regressive and hierarchy-2), with references to the FDM paper for the later}

%\ram{Mention how we made the comparison fair between different schemes}

To allow for fair comparison, all baseline schemes, including the standard autoregressive rollout and the \textit{hierarchy-2} scheme from the FDM model~\cite{harvey2022flexible}, were used with the same temporal and computational budget as our multiscale method. 
% Specifically, all models were limited to using templates whose total temporal span did not exceed that of our largest multiscale template, which covers $19$ time steps. 
All schemes are restricted to generating exactly three new frames at once, with four context frames, resulting in a fixed window size of \(2K+1 = 7\) frames (\(K = 3\)). This required adapting the hierarchy-2 baseline to limit the number of generated frames per call. 
In addition, all models are limited to generating a video which extends at maximum $9$ time steps in the future, and $9$ time steps in the past. Thus, each generated video spans at maximum $19=9+1+9$ steps (see for example our multiscale inference scheme in Fig.~\ref{fig:app-multiscale-inference}, horizon $9$).

\paragraph{Autoregressive scheme.}
The autoregressive baseline progresses through time using a fixed-length context window. At each step, it generates three future frames conditioned on the most recent four, repeating this process uniformly across the sequence. 
% This strategy relies entirely on short-term context and does lacks mechanisms for incorporating long-range temporal dependencies.

\paragraph{Hierarchy-2 scheme.}
The hierarchy-2 scheme, as described in the FDM framework~\cite{harvey2022flexible}, first generates a coarse sequence of future frames and then fills in the intermediate steps through recursive refinement. 
In our experiments, this scheme is adapted to output three frames per step while respecting the 19-frame context limit set above for fair comparisons. 
Despite this adjustment, the method retains its hierarchical trajectory structure, providing an alternative to the autoregressive rollouts.

\paragraph{Multiscale scheme (ours).}
Our proposed multiscale inference scheme relies on multiscale templates to capture long-range dependencies efficiently.
This design exploits the long-memory characteristics of physical systems like solar dynamics, enabling the model to integrate coarse and fine temporal context. The reasoning behind our multiscale template stems from the long-memory property inherent to such physical processes, as illustrated in Figure~\ref{fig:app-autocorrelation}. This phenomenon has been extensively studied in the context of scale-invariant processes~\cite{mccoy1996wavelet,abry2000wavelets,mandelbrot2013multifractals,morel2025scale}, particularly through the use of wavelet-based analysis~\cite{stephane1999wavelet}.
Once the inference strategy is fixed (see Fig.~\ref{fig:app-multiscale-inference}, with a horizon of 9), the corresponding templates, and masks defining the conditioning and generated data, are used during training.

\paragraph{Positional encodings.} 
To enable our model to generate videos with varying time steps, we incorporate temporal information at multiple stages of the denoiser. First, time indices of the frames are added as input channels. Second, in the ViT denoiser, attention layers incorporate relative time indices using a RoPE~\cite{rope} positional encoding. In addition to frame time positional encodings, we also add pixel-wise latitude and longitude as additional channels to all models, facilitating accurate predictions near the limb of the Sun.

\subsection{Baseline architectures}
%\ram{One paragraph for each of the baselines we took}
%\ram{Mention how we made the comparison fair between different models}
In addition to aligning inference schemes, we ensure that all model architectures are compared under identical training conditions. Each model is trained for 40 epochs on the same dataset, with consistent preprocessing, with similar hyperparameters (see below). In particular, all models have roughly $60$M parameters. 
% and a fixed context length of seven frames, as described in the previous section. 
% This uniform setup guarantees that any performance difference stems from architectural design rather than variation in training regime or data exposure.

\noindent\textbf{Axial ViT (AViT).}
The Axial Vision Transformer (AViT)~\cite{mccabe2024multiple} employs axial attention to model spatiotemporal dependencies efficiently in high-dimensional sequences. 
In our paper, an AViT is paired with the standard autoregressive inference scheme, it takes four frames as input (the conditioning) and predicts the three following frames. 
As a fully deterministic model, it serves as a strong baseline for assessing the value of stochastic modeling in solar forecasting.

\noindent\textbf{Auto-regressive diffusion (AR-diffusion).}
The AR-diffusion model~\cite{kohl2023benchmarking} applies diffusion sampling in a step-by-step autoregressive fashion. 
At each iteration of the sampling process, the model generates the next state conditioned on $4$ previous steps.

% \noindent\paragraph{Flexible Diffusion Model (FDM).} The FDM~\cite{harvey2022flexible} is trained using randomized conditioning masks that allow flexible inference at test time. In this work, we evaluate FDM using its hierarchy-2 sampling scheme, adapted to match our temporal generation constraints. This baseline enables coarse-to-fine inference and serves as a point of comparison for our multiscale rollout.

\subsection{Training hyperparameters}

We use the AdamW optimizer with a learning rate of \(1 \times 10^{-4}\), cosine learning rate scheduling, and a batch size of 64. Input patches are of shape \(1 \times 8 \times 8\), and the model consists of $16$ transformer blocks with $4$ attention heads each. 
All models are trained on $40$ epochs. Each epoch consists of $2000$ batches with a batch size of $64$, covering most of the training dataset.
All models were trained using a single node with 8 NVIDIA H100 GPUs.

\section{Multiscale inference schemes with longer horizons}
\label{app:longer-multiscale-schemes}

\begin{figure}[t]
\centering
\begin{subfigure}{1\textwidth}
  \centering
  \includegraphics[width=\textwidth]{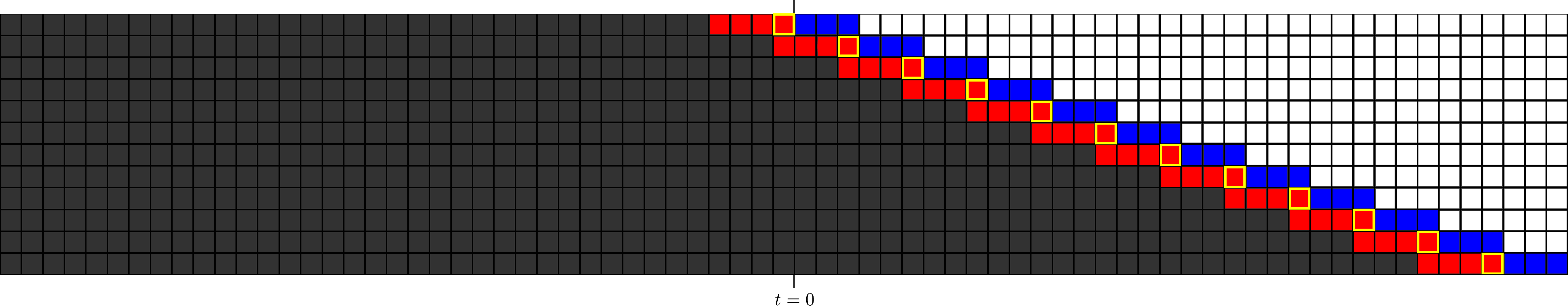}
  \caption*{Autoregressive}
\end{subfigure}
\vspace{0.3em}

\begin{subfigure}{\textwidth}
  \centering
  \includegraphics[width=\textwidth]{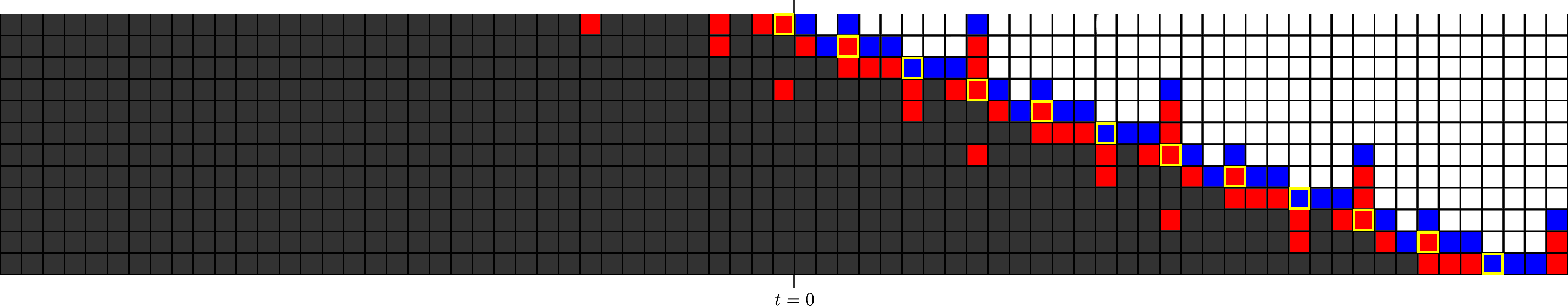}
  \caption*{Multiscale with horizon 9 (used in experiments)}
\end{subfigure}
\vspace{0.3em}

\begin{subfigure}{\textwidth}
  \centering
  \includegraphics[width=\textwidth]{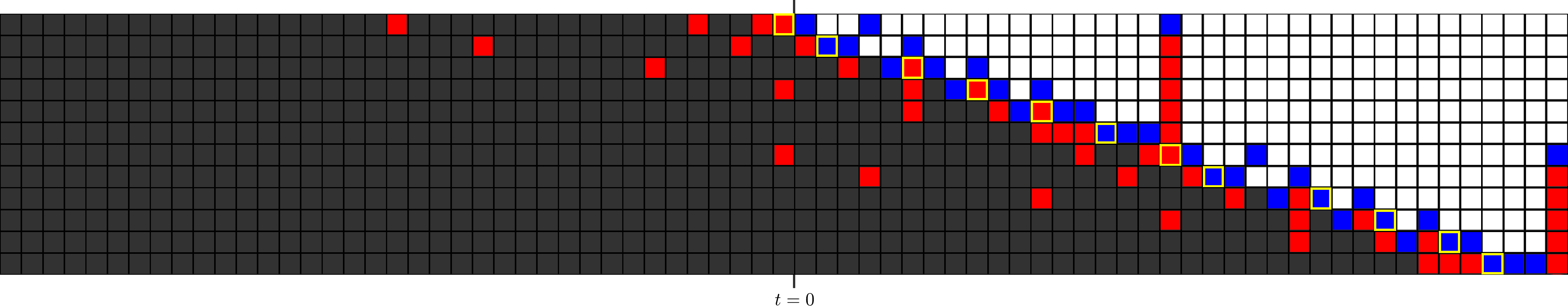}
  \caption*{Multiscale with horizon 18}
\end{subfigure}
\vspace{0.3em}

\begin{subfigure}{\textwidth}
  \centering
  \includegraphics[width=\textwidth]{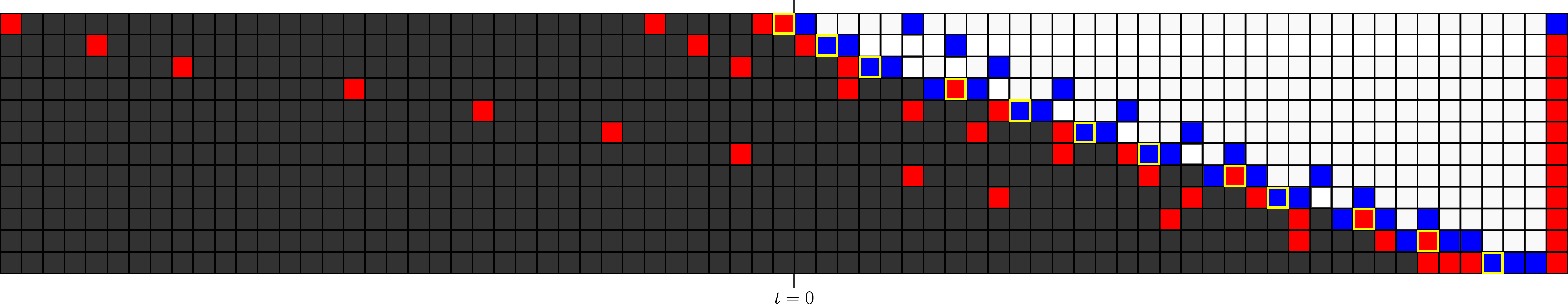}
  \caption*{Multiscale with horizon 36}
\end{subfigure}

\caption{
\textbf{Multiscale inference schemes with various horizons.}
The horizon denotes the furthest time step the model can predict in a single shot under a given inference scheme. For example, a horizon of $3$ corresponds to standard autoregressive prediction. Increasing the horizon allows the model to be conditioned more frequently on the past, including distant steps, enabling more stable long-range forecasting at the same inference cost.
}
\label{fig:app-multiscale-inference}
\end{figure}

In the experiments we chose multiscale templates of size $2K+1=7$ and a longest template $\T = \{-9,-3,-1,0,1,3,9\}$, with a horizon of $9$ steps in both past and future directions.
This inference scheme, shown in Fig.~\ref{fig:app-multiscale-inference}, was used in all experiments.

To better capture long-range dependencies, one can consider longer templates and more complex multiscale inference schemes. 
Fig.~\ref{fig:app-multiscale-inference} shows multiscale inference schemes based on a longest template $\T = \{-18, -4, -1, 0, 1, 4, 18\}$ with horizon $18$, and on a longest template $\T = \{-36, -6, -1, 0, 1, 6, 36\}$ with horizon $36$. Algorithm~\ref{alg:method-algo} presents a general inference scheme. 

\begin{figure}
\centering
\begin{algorithm}[H]
\caption{
Multiscale inference scheme for generic future horizon $H$, and time steps generated at once $K$. 
}
\label{alg:method-algo}
\begin{algorithmic}[1]
\STATE \textbf{Input:} Integer future horizon $H$, generation size $K$ (time steps), template set $\mathcal{T} = \{\tau^{(1)}, \dots, \tau^{(N)}\}$ ordered by increasing horizon
\STATE \textbf{Output:} Action list $A = [(n, \text{shift}, \text{mask})_1, \dots, (n, \text{shift}, \text{mask})_M]$ specifying template index, time shift, and conditioning mask at each generation step

\STATE Initialize: completed$[t] \leftarrow \text{False}$ for $t = 1, \dots, H$
\STATE Initialize: $A \leftarrow []$

\WHILE{$\exists t \in \{1, \dots, H\}$ with completed$[t] = \text{False}$}
    \STATE best $\leftarrow$ None, best\_score $\leftarrow \infty$
    \FOR{$n = N$ \TO $1$}
        \FOR{shift $= 0$ \TO $H$}
            \IF{$\max(\tau^{(n)}) + \text{shift} > H$}
                \STATE \textbf{continue}
            \ENDIF
            \STATE $I \leftarrow \{t + \text{shift} : t \in \tau^{(n)}\}$
            \STATE \textit{\# Check that the shifted template is conditioned on exactly $K+1$ steps}
            \STATE overlap $\leftarrow |\{t \in I : \text{completed}[t] = \text{True}\}|$
            \IF{overlap $= K+1$}
                \STATE $I_{\text{future}} \leftarrow \{t \in I : t > 0\}$
                \STATE $\mathcal{G} \leftarrow \{t \in \{1,\dots,H\} : \text{completed}[t] = \text{True}\}$
                \STATE $L \leftarrow |\mathcal{G} \setminus I_{\text{future}}|$  \textit{\# Steps already generated not covered by the template}
                \STATE $a \leftarrow (\max(\tau^{(n)}) + \text{shift} = H)$
                \STATE mask $\leftarrow [\text{completed}[t]]_{t \in I}$  \textit{\# Store which indices are conditioned}
                \STATE \textit{\# Favors a template which anchors on the very last future step}
                \IF{$a = \text{True}$}
                    \STATE best $\leftarrow (n, \text{shift}, \text{mask})$
                    \STATE \textbf{break} both loops
                \ELSE
                    \STATE \textit{\# Favors maximal self-conditioning for temporal coherence}
                    \IF{$L <$ best\_score}
                        \STATE best $\leftarrow (n, \text{shift}, \text{mask})$
                        \STATE best\_score $\leftarrow L$
                    \ENDIF
                \ENDIF
            \ENDIF
        \ENDFOR
    \ENDFOR
    
    \IF{best $\neq$ None}
        \STATE Unpack $(n_{\text{best}}, \text{shift}_{\text{best}}, \text{mask}_{\text{best}}) \leftarrow$ best
        \STATE $I_{\text{best}} \leftarrow \{t + \text{shift}_{\text{best}} : t \in \tau^{(n_{\text{best}})}, t > 0\}$
        \STATE Mark completed$[t] \leftarrow \text{True}$ for all $t \in I_{\text{best}}$
        \STATE Append $(n_{\text{best}}, \text{shift}_{\text{best}}, \text{mask}_{\text{best}})$ to $A$
    \ELSE
        \STATE \textbf{break}
    \ENDIF
\ENDWHILE

\RETURN $A$
\end{algorithmic}
\end{algorithm}
\end{figure}

\section{Additional model evaluation on solar dynamics}
\label{app:additional-eval}

\subsection{Physical parameters evaluation}

%\ram{Looking at the main text, we have: NMAE MeanGBT, NMAE MeanGBZ, they should be clearly presented (the reader must see in one look where are their definition), with references.}

We evaluate the physical validity of predictions using SHARP parameters (Table~\ref{tab:app-metrics-sharps}) commonly used in solar physics to characterize active regions and their flare potential~\cite{Bobra2015,LekaBarnes2003a,LekaBarnes2007}.
Just like the power spectrum, these statistics are computed on a single state.
% To ensure that our model captures a physically meaningful process, we evaluate its performance using metrics that are widely accepted within the solar physics community. Our objective is not only to generate visually plausible predictions, but to produce results that are scientifically interpretable and useful for solar research. To this end, we evaluate our predictions using a subset of parameters from the SHARP (Space-weather HMI Active Region Patch) data product (Table \ref{tab:app-metrics-sharps}). These parameters were introduced by \cite{Bobra2015} to quantify properties of solar active regions known to correlate with increased flare productivity \cite{LekaBarnes2003a, LekaBarnes2007}.

\paragraph{SHARPs computation.} We first map our vector magnetic field into the Cylindrical Equal-Area (CEA) system of reference. This is done because it is important to be in a uniform coordinate grid (equal area per pixel), while in native HMI CCD coordinates pixel scale varies across the field of view due to projection effects, which makes it invalid to integrate or compare pixel by pixel across the region. CEA corrects for that by projecting the data into a grid where each pixel covers the same surface area on the solar sphere.
While SHARP includes sixteen parameters in total, we focus on three representative and intuitive quantities: the total unsigned magnetic flux, the horizontal gradient of the total magnetic field, and the horizontal gradient of the vertical magnetic field. This selection allows us to analyze all three components of the vector magnetic field while maintaining interpretability and physical relevance.

\paragraph{Total unsigned flux (UsFlux).}
The total unsigned flux is computed from the radial component of the vector magnetic field, $|B_z|$, and represents the total absolute magnetic flux through the area $A$ of the active region:
\begin{equation}
\Phi = \int |B_z|\,dA \simeq \sum |B_z|\,dA.
\end{equation}
This quantity measures the amount of magnetic energy stored in the region and serves as a proxy for its magnetic complexity. The unsigned flux is usually directly proportional to the flaring probability of the region \cite{Li_2021}.

\paragraph{Horizontal Gradient of Total Field (MeanGBT).}
The horizontal gradient of the total magnetic field is defined as:
\begin{equation}
\overline{|\nabla B_{\text{tot}}|} = \frac{1}{N}\sum\sqrt{\left(\frac{\partial B}{\partial x}\right)^2 + \left(\frac{\partial B}{\partial y}\right)^2}, \quad \text{with} \quad B = \sqrt{B_x^2 + B_y^2 + B_z^2},
\end{equation}
and $N$ denoting the number of pixels over which the sum is computed.
This parameter quantifies how rapidly the total magnetic field strength changes across the horizontal plane. High values of this gradient indicate complex magnetic structures with strong shear or twist, which are typically associated with the onset of solar flares \cite{Li_2021}.

\paragraph{Horizontal Gradient of Vertical Field (MeanGBZ).}
The horizontal gradient of the radial (vertical) component $B_z$ is given by:
\begin{equation}
\overline{|\nabla B_z|} = \frac{1}{N}\sum\sqrt{\left(\frac{\partial B_z}{\partial x}\right)^2 + \left(\frac{\partial B_z}{\partial y}\right)^2},
\end{equation}
with $N$ the number of pixels over which the sum is computed.
This metric reflects how quickly the radial field changes in the horizontal direction. High values of $\overline{|\nabla B_z|}$ often appear near polarity inversion lines, where the magnetic field polarity reverses. These regions are important because they are commonly associated with magnetic reconnection events that can trigger solar flares \cite{Sharykin_2020}.

Tab.~\ref{tab:app-metrics-sharps} shows the 
% To quantitatively evaluate the quality of our model's predictions, we compute the 
Normalized Mean Absolute Error (NMAE) between predicted and observed values for each SHARP parameter. This metric is defined as the mean absolute error normalized by the mean absolute value of the observed parameter.
% , providing a scale-invariant measure of relative error. 
Specifically, for a given metric \( M \), the NMAE is defined as:
\begin{equation}
\nonumber
\text{NMAE}(M) = \frac{1}{N} \sum_{i=1}^{N} \frac{|M_i - M^\text{obs}_i|}{|M^\text{obs}_i|}\,, 
% \quad \text{where} \quad \overline{M} = \frac{1}{N} \sum_{i=1}^{N} M^{GT}_i
\end{equation}
where $M_i$ is computed on a predicted state, and $M^\text{obs}_i$ is computed on the observed state.
% In this case GT stands for ground truth and $M^{GT}$ represent the SHARP parameter computed on the original data, while PR stands for predicted and $M^{PR}$ represent the SHARP parameter computed on the predicted data.
% By applying NMAE to physically meaningful quantities, we ensure that our evaluation reflects not only visual similarity but also scientific accuracy with respect to key magnetic properties of solar active regions.

\begin{table}
\caption{
\textbf{Predictions performances measured with physical parameters.}
We compare different inference schemes (Autoregressive, Hierarchy-2~\cite{harvey2022flexible}, Ours – Multiscale). 
For each, we evaluate at three different time windows (1:4 hours, 4:16 hours, 16:32 hours) using multiple metrics: the normalized mean absolute error of representative solar physics quantities from~\cite{sharps} – the unsigned flux (UsFlux), the Mean Horizontal Gradient of the Total Field (MeanGBT) and of the Vertical Field (MeanGBZ)
% Sharps parameters \fpr{I guess this is NMAE too right?}
}
\label{tab:app-metrics-sharps}
\centering
\begin{tabular}{llllccc}
\toprule
\multicolumn{3}{l}{} & & \multicolumn{3}{c}{Relative error} \\
\cmidrule(r){1-3} \cmidrule(l){4-4} \cmidrule(l){5-7}
Model & Denoiser & Scheme & SHARP & 1:4 & 4:16 & 16:32 \\
\midrule
DiT & ViT & Autoreg. & UsFlux & 0.25 & 0.40 & 0.50 \\
DiT & ViT & Hierarchy-2~\cite{harvey2022flexible} & UsFlux & \textbf{0.19} & \textbf{0.38} & 0.52  \\
DiT & ViT & Multiscale (ours) & UsFlux & 0.23 & \textbf{0.38} & \textbf{0.48}  \\
\midrule
DiT & ViT & Autoreg. & MeanGBT & 0.18 & 0.30 & 0.37 \\
DiT & ViT & Hierarchy-2~\cite{harvey2022flexible} & MeanGBT & \textbf{0.12} & 0.28 & 0.38 \\
DiT & ViT & Multiscale (ours) & MeanGBT & 0.14 & \textbf{0.27} & \textbf{0.33} \\
\midrule
DiT & ViT & Autoreg. & MeanGBZ & 0.15 & 0.25 & 0.31 \\
DiT & ViT & Hierarchy-2~\cite{harvey2022flexible} & MeanGBZ & \textbf{0.088} & 0.22 & 0.31  \\
DiT & ViT & Multiscale (ours) & MeanGBZ & 0.10 & \textbf{0.21} & \textbf{0.27}  \\
\bottomrule
\end{tabular}
\end{table}

\subsection{Predicted trajectories}

\begin{figure}[t]
\centering
\includegraphics[width=\textwidth]{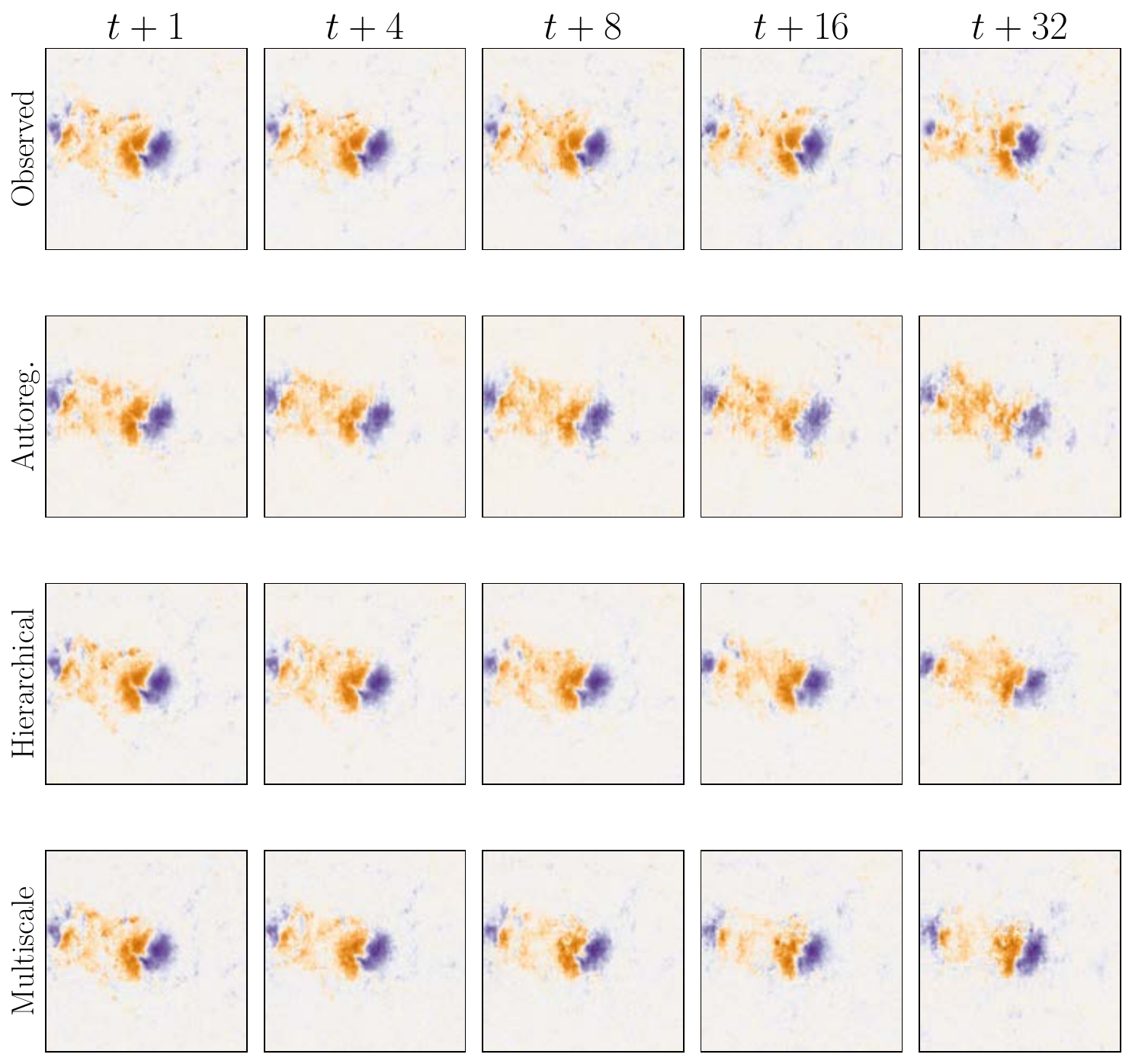}
\caption{
\textbf{Example of predictions (1/3), for different inference schemes.}
From top to bottom: observed data, autoregressive, hierarchy-2~\cite{harvey2022flexible}, multiscale (ours).
}
\label{fig:app-rollout-schemes-1}
\end{figure}

\begin{figure}[t]
\centering
\includegraphics[width=\textwidth]{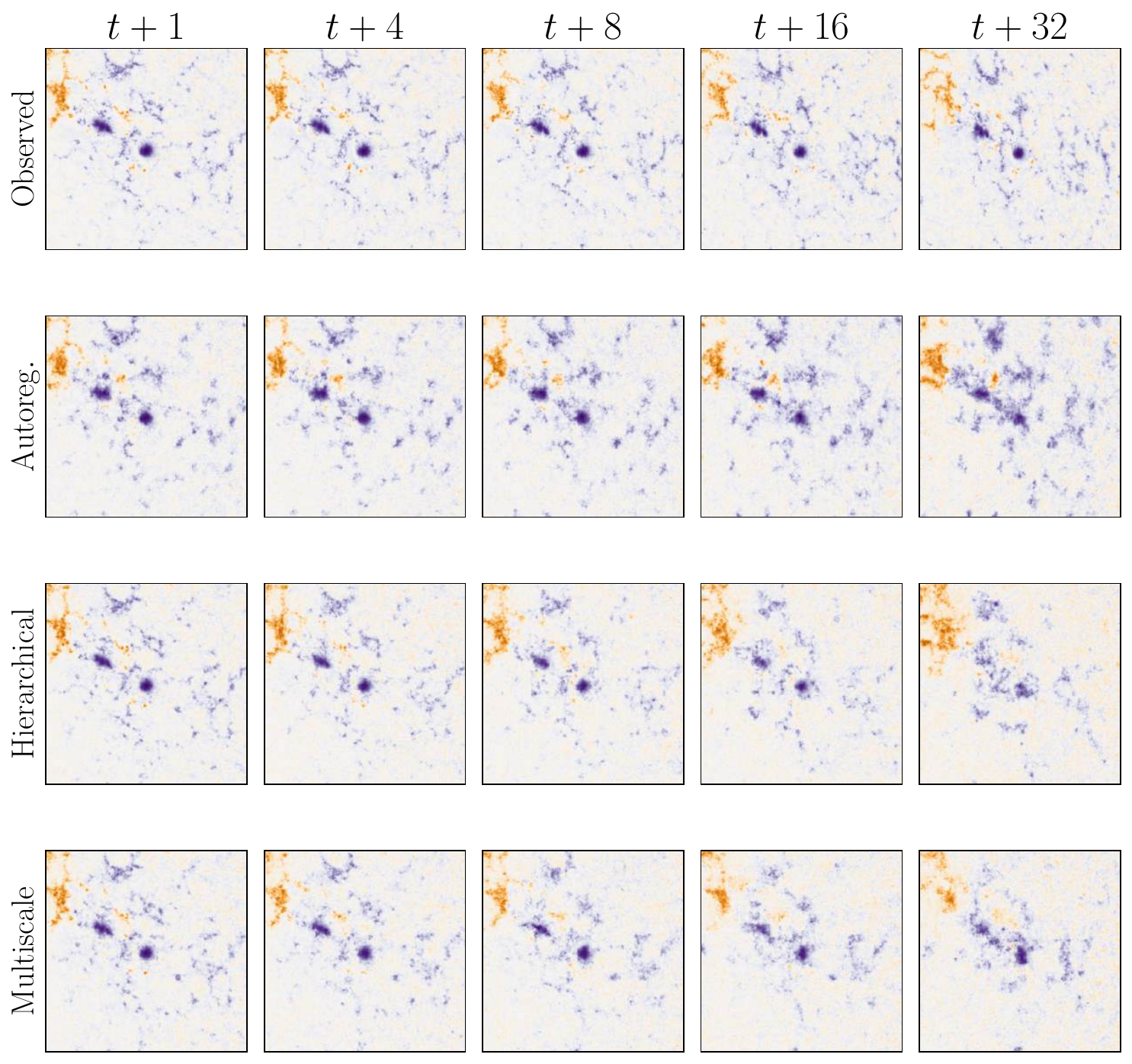}
\caption{
\textbf{Example of predictions (2/3), for different inference schemes.}
From top to bottom: observed data, autoregressive, hierarchy-2~\cite{harvey2022flexible}, multiscale (ours).
}
\label{fig:app-rollout-schemes-2}
\end{figure}

\begin{figure}[t]
\centering
\includegraphics[width=\textwidth]{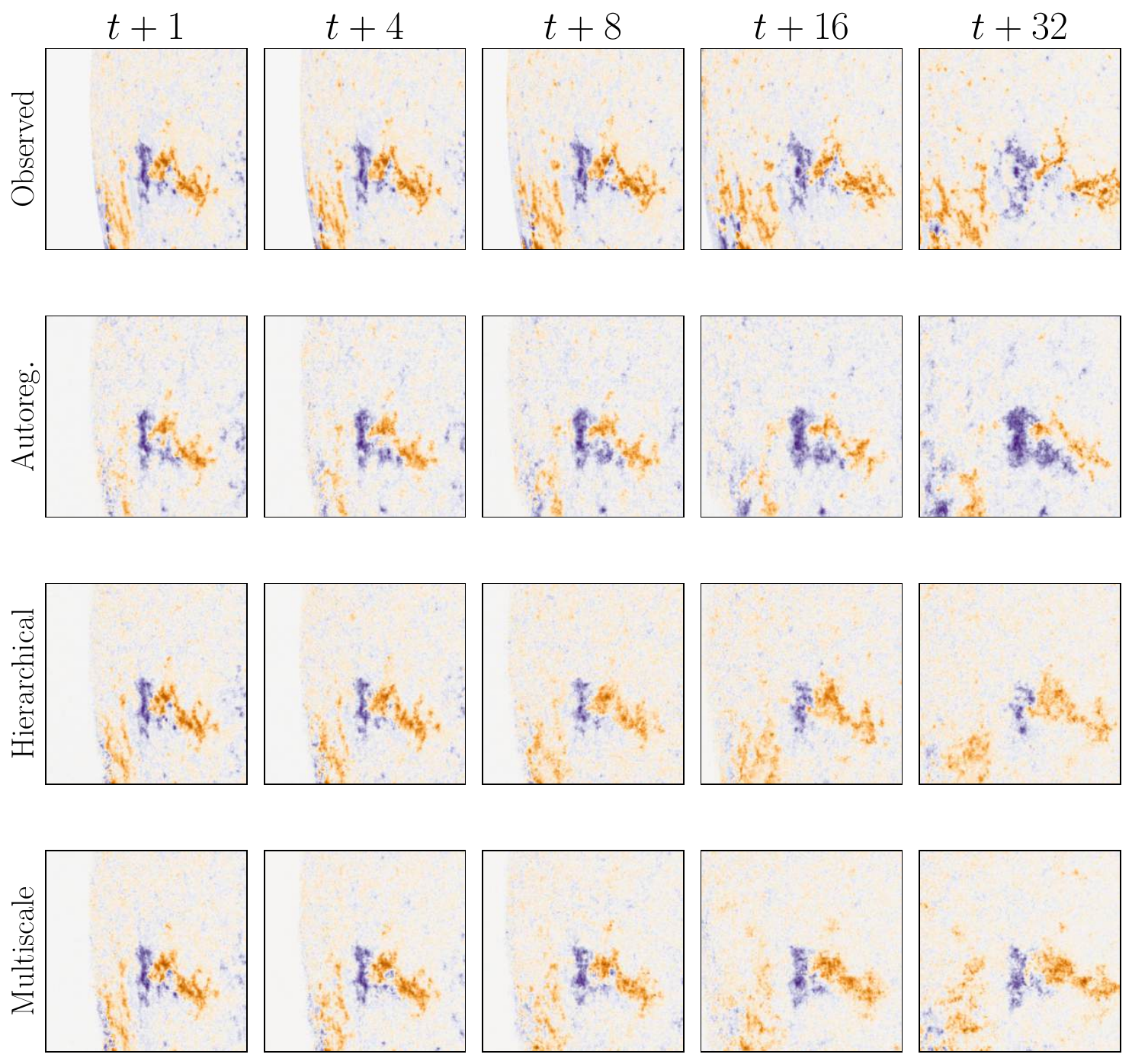}
\caption{
\textbf{Example of predictions (3/3), for different inference schemes.}
From top to bottom: observed data, autoregressive, hierarchy-2~\cite{harvey2022flexible}, multiscale (ours).
}
\label{fig:app-rollout-schemes-3}
\end{figure}

\begin{figure}[t]
\centering
\includegraphics[width=\textwidth]{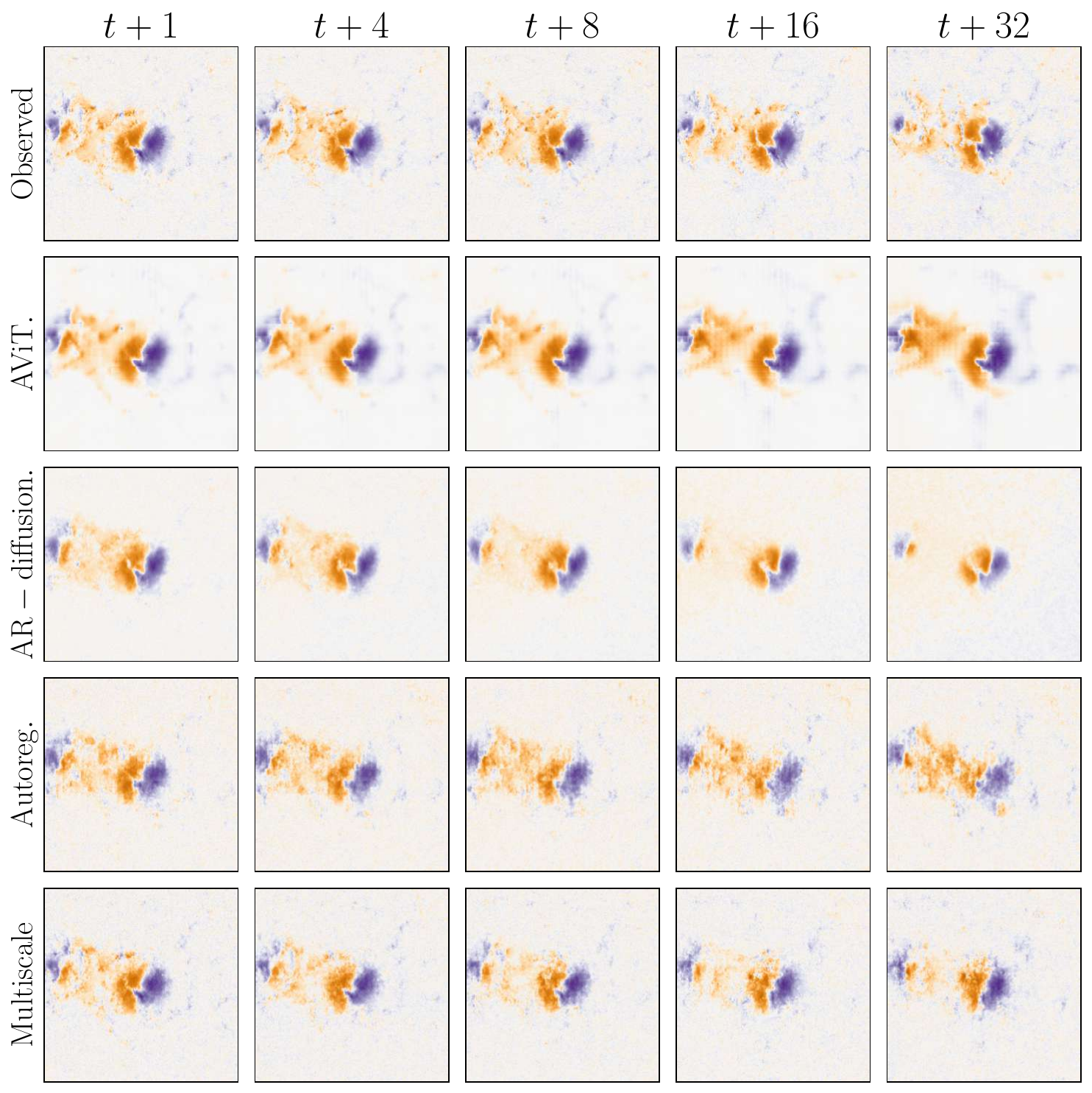}
\caption{
\textbf{Example of predictions (1/3), for different models.}
From top to bottom: observed data, AViT~\cite{mccabe2024multiple} model, autoregressive-diffusion model~\cite{kohl2023benchmarking}, our model with an autoregressive inference scheme, our model with a multiscale inference scheme.
}
\label{fig:app-rollout-model-1}
\end{figure}

\begin{figure}[t]
\centering
\includegraphics[width=\textwidth]{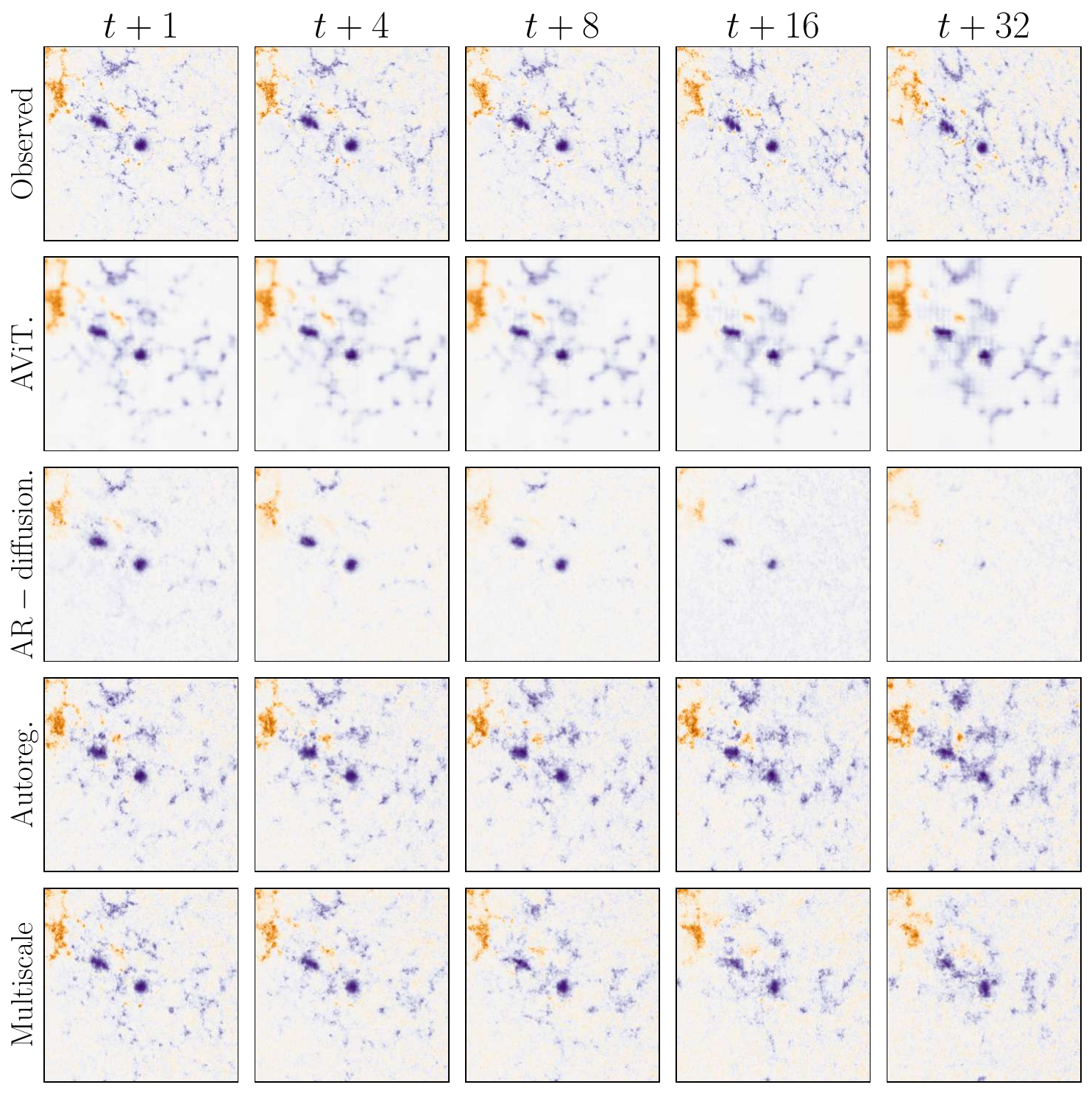}
\caption{
\textbf{Example of predictions (2/3), for different models.}
From top to bottom: observed data, AViT~\cite{mccabe2024multiple} model, autoregressive-diffusion model~\cite{kohl2023benchmarking}, our model with an autoregressive inference scheme, our model with a multiscale inference scheme.
}
\label{fig:app-rollout-model-2}
\end{figure}

\begin{figure}[t]
\centering
\includegraphics[width=\textwidth]{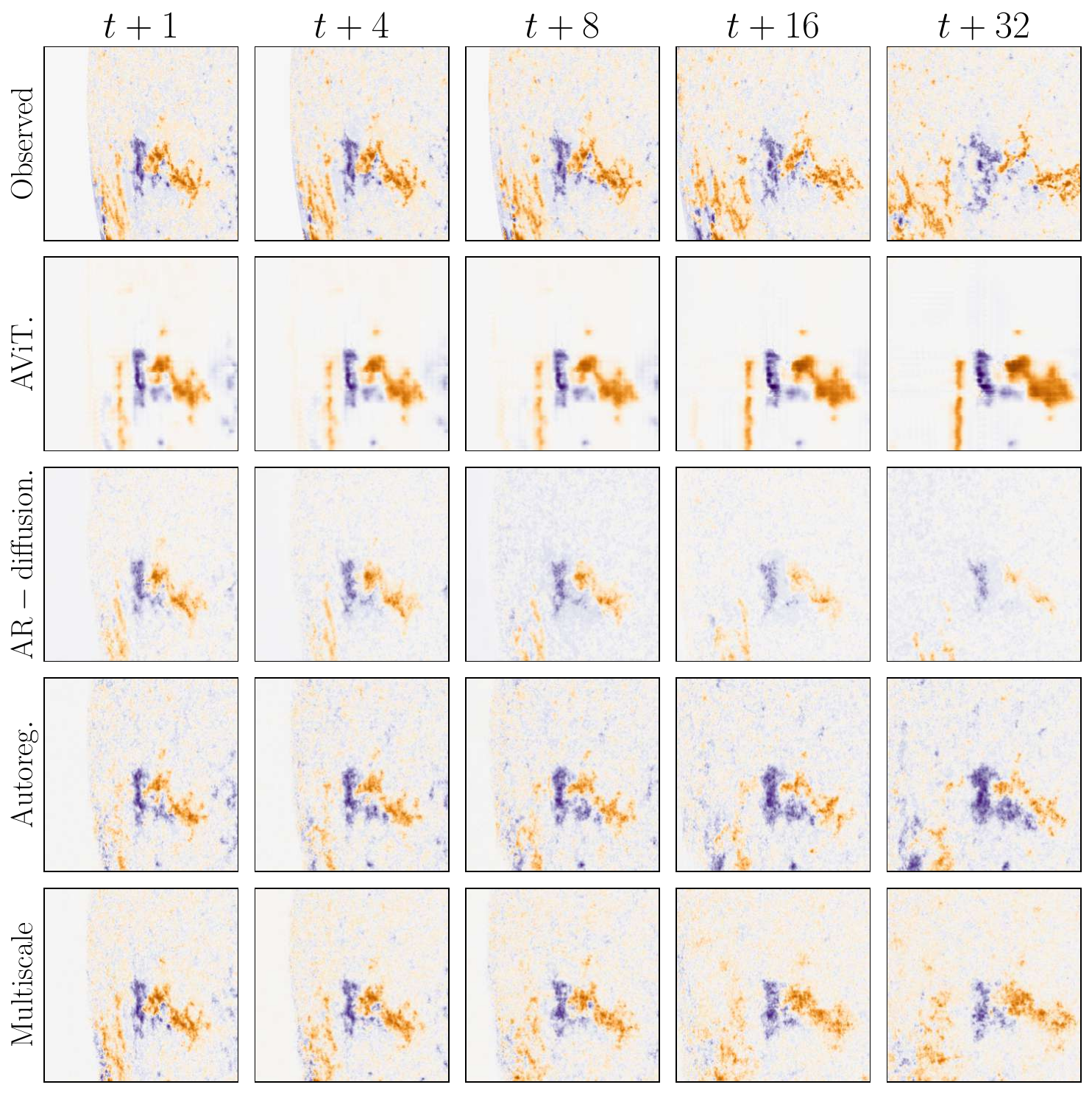}
\caption{
\textbf{Example of predictions (3/3), for different models.}
From top to bottom: observed data, AViT~\cite{mccabe2024multiple} model, autoregressive-diffusion model~\cite{kohl2023benchmarking}, our model with an autoregressive inference scheme, our model with a multiscale inference scheme.
}
\label{fig:app-rollout-model-3}
\end{figure}

Additional examples of predicted trajectories are shown for different inference schemes (Fig.~\ref{fig:app-rollout-schemes-1},\ref{fig:app-rollout-schemes-2},\ref{fig:app-rollout-schemes-3}) and for different models, with their respective preferred inference schemes (Fig.~\ref{fig:app-rollout-model-1},\ref{fig:app-rollout-model-2},\ref{fig:app-rollout-model-3}).

%%%%%

\section{Additional experiments on a fluid dynamics dataset}

To showcase the generality of our multiscale inference scheme, we applied it to the Navier-Stokes data from PDEArena~\cite{gupta2022towards}. 
% To make the dynamical system partially observable, we downsampled it to $32\times32$ (from $128\times128$) and considered only the density field (discarding velocity). 
To make the dynamical system partially observable, we first downsampled it temporally by a factor of $5$, then spatially by a factor of $4$ (from $128\times128$), and finally retained only the density field (discarding velocity).
The results on this new example of partially observable process are presented in Tab.~\ref{tab:synthetic-pdearena} below using the same metric as in the main paper (see MAE on the power-spectrum in Tab.~\ref{tab:main-comparison-table}).

\begin{table}[h!]
\caption{
{\bf Synthetic example: partially observable fluid dynamics.}
We compare different inference schemes (Autoregressive and Multiscale -- ours) on a synthetic fluid dyanmics example, evaluated over three time windows: 1:4, 4:16, and 16:36 steps. The metric is the Mean Absolute Error (MAE) of the power spectrum.
}
\vspace{0.2cm}
\label{tab:synthetic-pdearena}
\centering
\setlength{\tabcolsep}{0.37em}
\newcommand{\smallrange}{{\small 1:4}}
\newcommand{\midrange}{{\small 4:16}}
\newcommand{\bigrange}{{\small 16:36}}
\begin{tabular}{@{}lcccc@{}}
\toprule
\multicolumn{1}{l}{} & \multicolumn{3}{c}{MAE Power Spectrum} \\
\cmidrule(lr){2-4}
Inference scheme & \smallrange & \midrange & \bigrange \\
\midrule
Autoregressive    & 0.39 & 0.43 & 0.35 \\
Multiscale (ours) & \textbf{0.38} & \textbf{0.36} & \textbf{0.31} \\
\bottomrule
\end{tabular}
\end{table}

We observe that our multiscale inference scheme outperforms the autoregressive baseline, particularly in the long-term intervals (4:16 and 16:36), where it achieves a significantly lower MAE on the power spectrum. This demonstrates that our approach is not specific to solar dynamics prediction but can be applied successfully to other partially observable systems.

% \bibliographystyle{plain}  % or unsrt, ieee, etc.
% \bibliography{biblio}

\end{document}